\documentclass[sigconf, screen]{acmart}

\usepackage{graphicx}
\usepackage{amsmath}
\usepackage{booktabs}
\usepackage{makecell}
\usepackage{multirow}

\AtBeginDocument{%
  \providecommand\BibTeX{{%
    \normalfont B\kern-0.5em{\scshape i\kern-0.25em b}\kern-0.8em\TeX}}}

\copyrightyear{2022} 
\acmYear{2022} 
\setcopyright{acmcopyright}\acmConference[MM '22]{Proceedings of the 30th ACM International Conference on Multimedia}{October 10--14, 2022}{Lisboa, Portugal}
\acmBooktitle{Proceedings of the 30th ACM International Conference on Multimedia (MM '22), October 10--14, 2022, Lisboa, Portugal}
\acmPrice{15.00}
\acmDOI{10.1145/3503161.3547808}
\acmISBN{978-1-4503-9203-7/22/10}

\begin{document}

\title{NeRF-SR: High Quality Neural Radiance Fields using Supersampling}

\author{Chen Wang}
\orcid{0000-0002-9315-3780}
\affiliation{%
  \institution{BNRist, Department of Computer Science and Technology, Tsinghua University}
  \city{Beijing}
  \country{China}
}

\author{Xian Wu}
\affiliation{%
  \institution{Kuaishou Technology}
  \city{Beijing}
  \country{China}
}

\author{Yuan-Chen Guo}
\affiliation{%
  \institution{BNRist, Department of Computer Science and Technology, Tsinghua University}
  \city{Beijing}
  \country{China}
}

\author{Song-Hai Zhang}
\orcid{0000-0003-0460-1586}
\affiliation{%
 \institution{BNRist, Department of Computer Science and Technology, Tsinghua University}
 \city{Beijing}
 \country{China}
}

\author{Yu-Wing Tai}
\authornote{Corresponding author}
\affiliation{%
  \institution{Kuaishou Technology}
  \city{Beijing}
  \country{China}
}
\email{yuwing@gmail.com}

\author{Shi-Min Hu}
\orcid{0000-0001-7507-6542}
\affiliation{%
  \institution{BNRist, Department of Computer Science and Technology, Tsinghua University}
  \city{Beijing}
  \country{China}
}



\renewcommand{\shortauthors}{Wang, et al.}



\def\etal{\textit{et~al.~}}		  
\def\eg{e.g.,~}               
\def\ie{\textit{i.e.,~}}      
\def\etc{etc}                 
\def\cf{cf.~}                 
\def\viz{viz.~}               
\def\vs{vs.~}                 

\def\sysname{NeRF-SR}
\def\weburl{https://cwchenwang.github.io/NeRF-SR}
\def\suppurl{https://cwchenwang.github.io/NeRF-SR/data/supp.pdf}

\def\naive{na{\"i}ve\xspace}
\def\Naive{Na{\"i}ve\xspace}
\def\Naively{Na{\"i}vely\xspace}



\newlength\paramargin
\newlength\figmargin

\newlength\secmargin
\newlength\figcapmargin
\newlength\tabcapmargin

\setlength{\secmargin}{0.0mm}
\setlength{\paramargin}{0.0mm}
\setlength{\figmargin}{0.0mm}
\setlength{\tabcapmargin}{0.0mm}

\setlength{\figcapmargin}{1.0mm}

\setlength{\fboxsep}{0pt}

\newcommand{\red}{\textcolor{red}}
\newcommand{\blue}{\textcolor{blue}}

\newcommand{\mpage}[2]
{
\begin{minipage}{#1\linewidth}\centering
#2
\end{minipage}
}

\newcommand{\mfigure}[2]
{
\includegraphics[width=#1\linewidth]{#2}
}

\newcommand{\topic}[1]
{
\vspace{1.5mm}\noindent\textbf{#1}
}

\newcommand{\secref}[1]{Section~\ref{#1}}
\newcommand{\figref}[1]{Figure~\ref{#1}} 
\newcommand{\tabref}[1]{Table~\ref{#1}}
\newcommand{\eqnref}[1]{Equation~\eqref{#1}}
\newcommand{\thmref}[1]{Theorem~\ref{#1}}
\newcommand{\prgref}[1]{Program~\ref{#1}}
\newcommand{\algref}[1]{Algorithm~\ref{#1}}
\newcommand{\clmref}[1]{Claim~\ref{#1}}
\newcommand{\lemref}[1]{Lemma~\ref{#1}}
\newcommand{\ptyref}[1]{Property~\ref{#1}}

\long\def\ignorethis#1{}
\newcommand {\xxx}[1]{{\color{cyan}\textbf{: }#1}\normalfont}
\newcommand {\xxxx}[1]{{\color{red}\textbf{: }#1}\normalfont}
\newcommand {\xx}[1]{{\color{magenta}\textbf{: }#1}\normalfont}
\newcommand {\chen}[1]{{\color{blue}\textbf{Chen: }#1}\normalfont}

\newcommand {\todo}[1]{{\textbf{\color{red}[TO-DO: #1]}}}
\def\newtext#1{\textcolor{blue}{#1}}
\def\modtext#1{\textcolor{red}{#1}}
\newcommand{\note}[1]{{\it\color{blue} #1}}

\newcommand{\tb}[1]{\textbf{#1}}
\newcommand{\mb}[1]{\mathbf{#1}}

\newcommand{\jbox}[2]{
  \fbox{%
  	\begin{minipage}{#1}%
  		\hfill\vspace{#2}%
  	\end{minipage}%
  }}

\newcommand{\jblock}[2]{%
	\begin{minipage}[t]{#1}\vspace{0cm}\centering%
	#2%
	\end{minipage}%
}

\newbox\jsavebox%
\newcommand{\jsubfig}[2]{%
	\sbox\jsavebox{#1}%
	\parbox[t]{\wd\jsavebox}{\centering\usebox\jsavebox\\#2}%
	}

\makeatletter
\newcommand{\providelength}[1]{%
  \@ifundefined{\expandafter\@gobble\string#1}
   {
    \typeout{\string\providelength: making new length \string#1}%
    \newlength{#1}%
   }
   {
    \sdaau@checkforlength{#1}%
   }%
}


\begin{abstract}
We present \sysname{}, a solution for high-resolution (HR) novel view synthesis with mostly low-resolution (LR) inputs. Our method is built upon Neural Radiance Fields (NeRF) \cite{mildenhall2020nerf} that predicts per-point density and color with a multi-layer perceptron. While producing images at arbitrary scales, NeRF struggles with resolutions that go beyond observed images. Our key insight is that NeRF benefits from 3D consistency, which means an observed pixel absorbs information from nearby views. We first exploit it by a supersampling strategy that shoots multiple rays at each image pixel, which further enforces multi-view constraint at a sub-pixel level. Then, we show that \sysname{} can further boost the performance of supersampling by a refinement network that leverages the estimated depth at hand to hallucinate details from related patches on only one HR reference image. Experiment results demonstrate that \sysname{} generates high-quality results for novel view synthesis at HR on both synthetic and real-world datasets without any external information. Project page: \href{https://cwchenwang.github.io/NeRF-SR}{https://cwchenwang.github.io/NeRF-SR}
\end{abstract}

\ccsdesc[500]{Computing methodologies~Computer vision}
\ccsdesc[500]{Computing methodologies~Computer graphics}
\keywords{Neural Radiance Fields, Super-resolution}


\maketitle

\section{Introduction}
\label{sec:intro}
Synthesizing photorealistic views from a novel viewpoint given a set of posed images, known as \textit{novel view synthesis}, has been a long-standing problem in the computer vision community, and an important technique for VR and AR applications such as navigation, and telepresence. Traditional approaches mainly falls in the range of image-based rendering and follows the process of warping and blending source frames to target views \cite{gortler1996lumigraph, levoy1996light}. Image-based rendering methods heavily rely on the quality of input data and only produces reasonable renderings with dense observed views and accurate proxy geometry.

\begin{figure}[htbp]
    \centering
    \includegraphics[width=\linewidth]{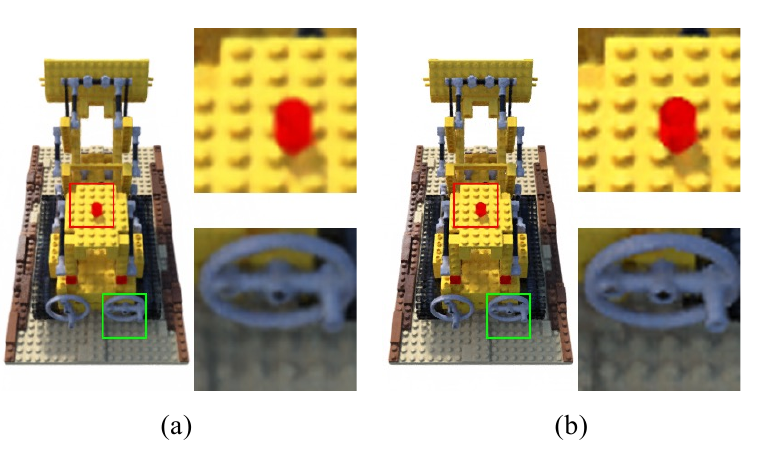}
    \caption{NeRF, the state-of-the-art novel view synthesis method, can synthesize photorealistic outputs at the resolution of training images but struggles at higher resolutions as shown in (a), while \sysname{} produces high-quality novel views (b) even with low-resolution inputs.}
    \label{fig:teaser}
\end{figure}

Most recently, \textit{neural rendering} has made significant progress on novel view synthesis by leveraging learnable components with 3D geometry context to reconstruct novel views with respect to input images. As the current state-of-the-art method, neural radiance fields (NeRF) \cite{mildenhall2020nerf} have emerged as a promising direction for neural scene representation even on sparse image sets of complex real-world scenes. NeRF uses the weights of multi-layer perceptrons (MLPs) to encode the radiance field and volume density of a scene. Most importantly, the implicit neural representation is continuous, which enables NeRF to take as input any position in the volume at inference time and render images at any arbitrary resolution.

At the same time, a high-resolution 3D scene is essential for many real-world applications, \eg a prerequisite to providing an immersive virtual environment in VR. However, a trained NeRF struggles to generalize directly to resolutions higher than that of the input images and generates blurry views (See \figref{fig:teaser}), which presents an obstacle for real-world scenarios, \eg images collected from the Internet may be low-resolution. To tackle this problem, we present \sysname{}, a technique that extends NeRF and creates high-resolution (HR) novel views with better quality even with low-resolution (LR) inputs. We first observe there is a sampling gap between the training and testing phase for super-resolving a 3D scene, since the sparse inputs are far from satisfying Nyquist view sampling rates~\cite{mildenhall2019local}. To this end, we derive inspiration from traditional graphics pipeline and propose a supersampling strategy to better enforce the multi-view consistency embedded in NeRF in a sub-pixel manner, enabling the generation of both SR images and SR depth maps. Second, in the case of limited HR images such as panoramas and light field imaging systems that have a trade-off between angular and spatial resolutions, we find that directly incorporating them in the NeRF training only improves renderings \textit{nearby the HR images} in a small margin. Thus, we propose a patch-wise warp-and-refine strategy that utilizes the estimated 3D geometry and propagate the details of HR reference to \textit{all over the scene}. Moreover, the refinement stage is efficient and introduces negligible running time compared with NeRF rendering.

To the best of our knowledge, we are the first to produce visually pleasing results for novel view synthesis under mainly low-resolution inputs. Our method requires only posed multi-view images of the target scene, from which we dig into the internal statistics and does not rely on any external priors. We show that \sysname{} outperforms baselines that require LR-HR pairs for training.

Our contributions are summarized as follows:
\begin{itemize}
    \item the first framework that produces decent multi-view super-resolution results with mostly LR input images
    \item a supersampling strategy that exploits the view consistency in images and supervises NeRF in the sub-pixel manner
    \item a refinement network that blends details from any HR reference by finding relevant patches with available depth maps
\end{itemize} 

\section{Related Work}
\label{sec:related-work}

\noindent\textbf{Novel View Synthesis.}
 Novel view synthesis can be categorized into image-based, learning-based, and geometry-based methods. Image-based methods warp and blend relevant patches in the observation frames to generate novel views based on measurements of quality \cite{gortler1996lumigraph, levoy1996light}. Learning-based methods predict blending weights and view-dependent effects via neural networks and/or other hand-crafted heuristics\cite{hedman2018deep, choi2019extreme, riegler2020free, thies2020image}. Deep learning has also facilitated methods that can predict novel views from a single image, but they often require a large amount of data for training\cite{tucker2020single, wiles2020synsin, shih20203d, niklaus20193d, rockwell2021pixelsynth}. Different from image-based and learning-based methods, geometry-based methods first reconstruct a 3D model \cite{schonberger2016structure} and render images from target poses. For example, Aliev \etal\cite{aliev2020neural} assigned multi-resolution features to point clouds and then performed neural rendering, Thies \etal\cite{thies2019deferred} stored neural textures on 3D meshes and then render the novel view with traditional graphics pipeline. Other geometry representations include multi-planes images \cite{zhou2018stereo, mildenhall2019local, flynn2019deepview, srinivasan2019pushing, li2020crowdsampling, li2021mine}, voxel grids \cite{henzler2020learning, penner2017soft, kalantari2016learning}, depth \cite{wiles2020synsin, flynn2019deepview, riegler2020free, riegler2021stable} and layered depth \cite{shih20203d, tulsiani2018layer}. These methods, although producing relatively high-quality results, the discrete representations require abundant data and memory and the rendered resolutions are also limited by the accuracy of reconstructed geometry.

\vspace{2mm}
\noindent\textbf{Neural Radiance Fields.} Implicit neural representation has demonstrated its effectiveness to represent shapes and scenes, which usually leverages multi-layer perceptrons (MLPs) to encode signed distance fields \cite{park2019deepsdf, duan2020curriculum}, occupancy \cite{mescheder2019occupancy, peng2020convolutional, chen2019learning} or volume density \cite{mildenhall2020nerf, niemeyer2020differentiable}. Together with differentiable rendering \cite{kato2018neural, liu2019soft}, these methods can reconstruct both geometry and appearance of objects and scenes \cite{sitzmann2019scene, saito2019pifu, niemeyer2020differentiable, sitzmann2019deepvoxels, liu2020neural}.  Among them, Neural Radiance Fields (NeRF) \cite{mildenhall2020nerf} achieved remarkable results for synthesizing novel views of a static scene given a set of posed input images. There are a growing number of NeRF extensions emerged, \eg reconstruction without input camera poses\cite{wang2021nerf, lin2021barf}, modelling non-rigid scenes \cite{pumarola2021d, park2021nerfies, park2021hypernerf, martin2021nerf}, unbounded scenes\cite{zhang2020nerf++} and object categories \cite{yu2021pixelnerf, trevithick2021grf, jang2021codenerf}. Relevant to our work, Mip-NeRF~\cite{barron2021mip} also considers the issue of \textit{resolution} in NeRF. They showed that NeRFs rendered at various resolutions would introduce aliasing artifacts and resolved it by proposing an integrated positional encoding that featurize conical frustums instead of single points. Yet, Mip-NeRF only considers rendering with downsampled resolutions. To our knowledge, no prior work studies how to increase the resolution of NeRF.


\vspace{2mm}
\noindent\textbf{Image Super-Resolution}
Our work is also related to image super-resolution. Classical approaches in single-image super-resolution (SISR) utilize priors such as image statistics \cite{kim2010single, zontak2011internal} or gradients \cite{sun2008image}. CNN-based methods aim to learn the relationship between HR and LR images in CNN by minimizing the mean-square errors between SR images and ground truths \cite{dong2014learning, wang2015deep, dong2015image}. Generative Adversarial Networks (GANs) \cite{goodfellow2014generative} are also popular in super-resolution which hallucinates high resolution details by adversarial learning \cite{ledig2017photo, menon2020pulse, sajjadi2017enhancenet}. These methods mostly gain knowledge from large-scale datasets or existing HR and LR pairs for training. Besides, these 2D image-based methods, especially GAN-based methods do not take the view consistency into consideration and are sub-optimal for novel view synthesis.

Reference-based image super-resolution (Ref-SR) upscales input images with additional reference high-resolution (HR) images. Existing methods match the correspondences between HR references and LR inputs with patch-match \cite{zhang2019image, zheng2017combining}, feature extraction \cite{xie2020feature, yang2020learning} or attention \cite{yang2020learning}. Although we also aim to learn HR details from given reference images, we work in the 3D geometry perspective and can bring details for all novel views instead of one image.
\section{Background}
\label{sec:background}
Neural Radiance Fields (NeRF) \cite{mildenhall2020nerf} encodes a 3D scene as a continuous function which takes as input 3D position $\mathbf{x} = (x, y, z)$ and observed viewing direction $\mathbf{d} = (\boldsymbol{\theta}, \boldsymbol{\phi})$, and predicts the radiance $\mathbf{c}(\mathbf{x}, \mathbf{d}) = (r, g, b)$ and volume density $\sigma(\mathbf{x})$. The color depends both on viewing direction $\mathbf{d}$ and $\mathbf{x}$ to capture view dependent effects, while the density only depends on $\mathbf{x}$ to maintain view consistency. NeRF is typically parametrized by a multilayer perceptron (MLP) $f: (\mathbf{x}, \mathbf{d}) \rightarrow (\mathbf{c}, \sigma)$.

NeRF is an emission-only model (the color of a pixel only depends on the radiance along a ray with no other lighting factors). Therefore, according to volume rendering \cite{kajiya1984ray}, the color along the camera ray $\mathbf{r}(t) = \mathbf{o} + t\mathbf{d}$ that shots from the camera center $\mathbf{o}$ in direction $\mathbf{d}$ can be calculated as:
\begin{equation}
    \mathbf{C}(\mathbf{r}) = \int_{t_n}^{t_f}T(t)\sigma(\mathbf{r}(t))\mathbf{c}(\mathbf{r}(t), \mathbf{d}) \mathrm{d}t
\label{equ:render}
\end{equation}
where
\begin{equation}
    T(t) = \mathrm{exp}\Big(-\int_{t_n}^{t}\sigma(\mathbf{r}(t))\mathrm{d}t\Big)
\end{equation}
is the accumulated transmittance that indicates the probability that a ray travels from $t_n$ to $t$ without hitting any particle.

NeRF is trained to minimize the mean-squared error (MSE) between the predicted renderings and the corresponding ground-truth color:
\begin{equation}
    \mathcal{L}_{\mathrm{MSE}} = \sum_{\mathbf{p} \in \mathcal{P}}\| \hat{\mathbf{C}}(\mathbf{r}_{\mathbf{p}}) - \mathbf{C}(\mathbf{r}_{\mathbf{p}}) \|_2^{2}
    \label{equ:mse}
\end{equation}
where $\mathcal{P}$ denotes all pixels of training set images, $\mathbf{r}_{\mathbf{p}}(t) = \mathbf{o} + t\mathbf{d}_{\mathbf{p}}$ denotes the ray shooting from camera center to the corners (or centers in some variants \cite{barron2021mip}) of a given pixel $\mathbf{p}$. $\hat{\mathbf{C}}(\mathbf{r}_{\mathbf{p}})$ and $\mathbf{C}(\mathbf{r}_{\mathbf{p}})$ are the ground truth and output color of $\mathbf{p}$. 

In practice, the integral in \eqnref{equ:render} is approximated by numeric quadrature that samples a finite number of points along with the rays and computes the summation of radiances according to the estimated per-point transmittance. The sampling in NeRF follows a \textit{coarse-to-fine} mechanism with two MLPs, \ie coarse network is queried on equally spaced samples whose outputs are utilized to sample another group of points for more accurate estimation and fine network is then queried on both groups of samples.

\section{Approach}
\label{sec:approach}
\begin{figure}
    \begin{center}
      \includegraphics[width=1.0\linewidth]{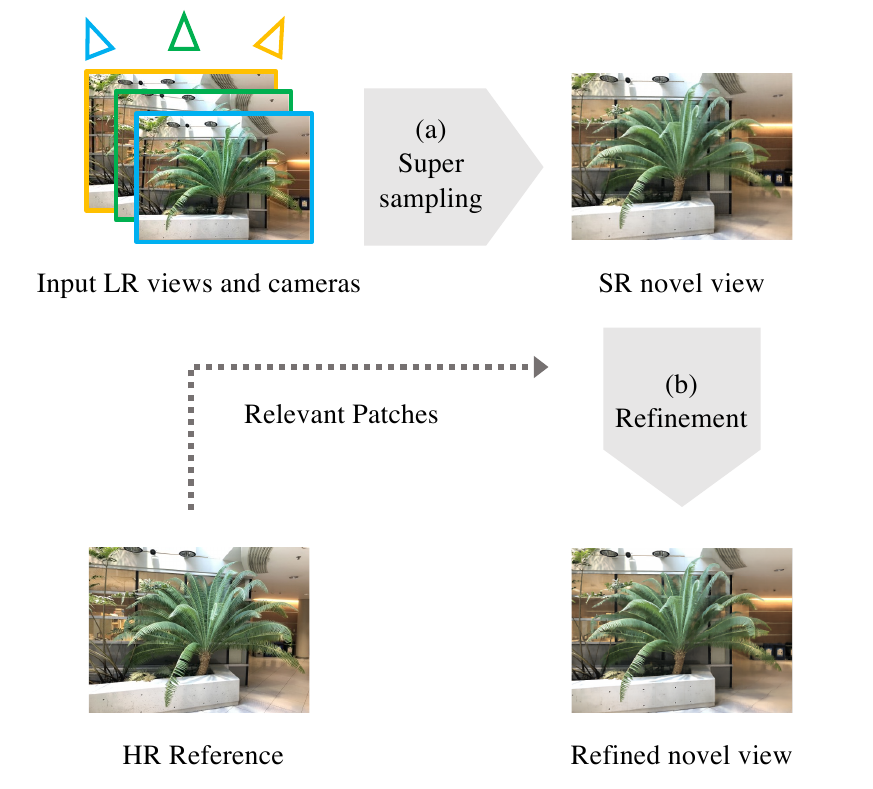}
    \end{center}
    \caption{An overview of the proposed \sysname{} that includes two components. (a), we adopt a super sampling strategy to produce super-resolution novel views from only low-resolution inputs. (b) Given an high-resolution reference at any viewpoint from which we utilize the depth map at hand to extract relevant patches, \sysname{} generates more details for synthesized images.}
    \label{fig:framework}
\end{figure}

In this section, we introduce the details of \sysname{}. The overall structure is presented in \figref{fig:framework}. The supersampling strategy and patch refinement network will be introduced in \secref{subsec:ss} and \secref{subsec:refine}.

\subsection{Supersampling}
\label{subsec:ss}
\begin{figure}
    \begin{center}
      \includegraphics[width=0.9\linewidth]{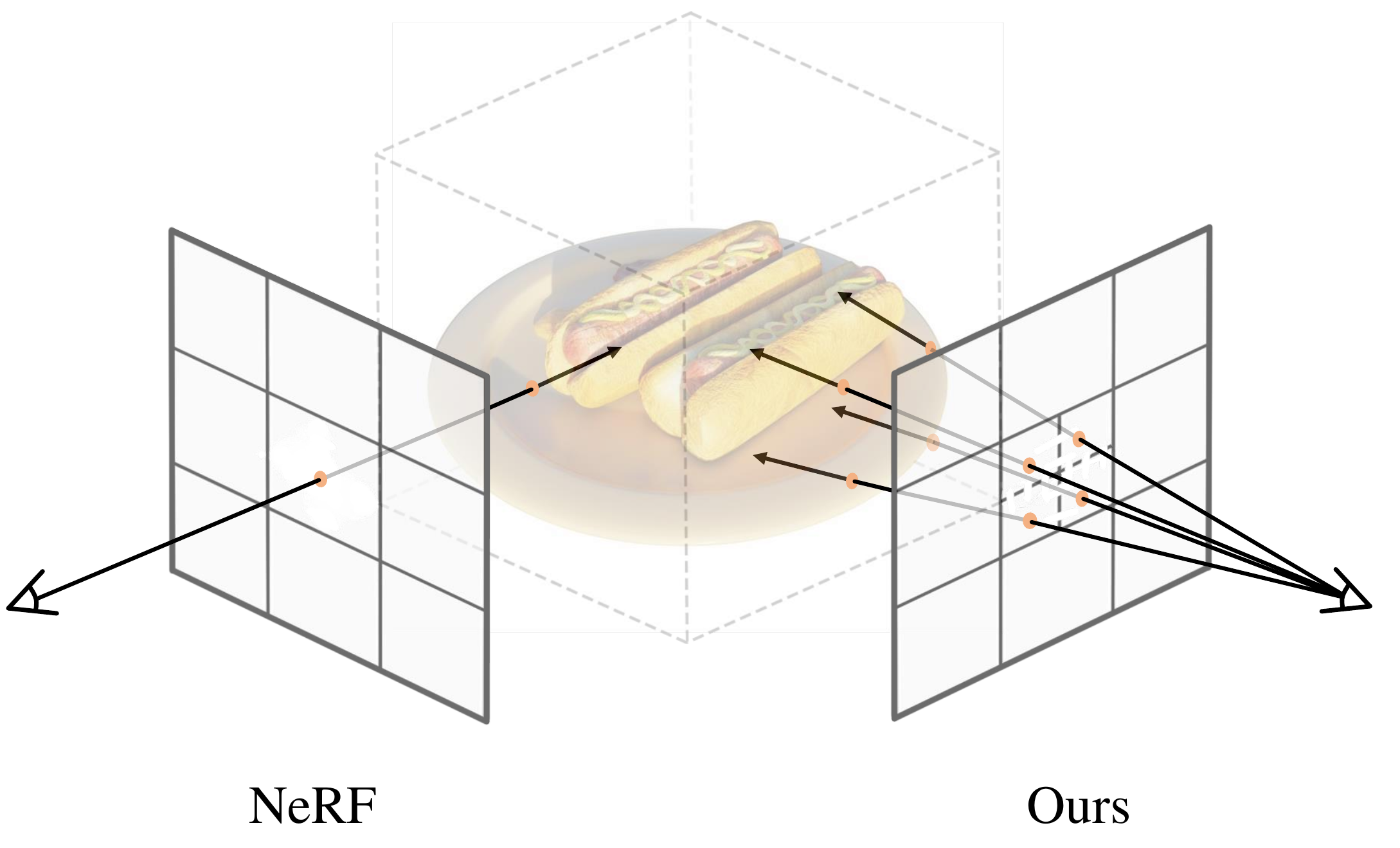}
    \end{center}
    \caption{Original NeRF casts a single ray through a pixel (solid line) and performance MSE loss directly (left), while our method (right) splits a pixel into multiple sub-pixels (dash line) and draws a ray for each sub-pixel, then the radiances of sub-pixels will be averaged for MSE loss. Compared to vanilla NeRF, more 3D points in the scene can be corresponded and constrained in supersampling.}
    \label{fig:super-sampling}
\end{figure}

NeRF optimizes a 3D radiance field by enforcing multi-view color consistency and samples rays based on camera poses and pixels locations in the training set. Although NeRF can be rendered at any resolution and retain great performance when the input images satisfy the Nyquist sampling rate, it is impossible in practice. Compared to the infinity possible incoming ray directions in the space, the sampling is quite sparse given limited input image observations. NeRF can create plausible novel views because the output resolution is the same as the input one and it relies on the interpolation property of neural networks. However, this becomes a problem when we render an image at a higher resolution than training images, specifically, there is a gap between the training and testing phase. Suppose a NeRF was trained on images of resolution $\mathrm{H} \times \mathrm{W}$, the most straightforward way to reconstruct a training image on scale factor $s$, \ie an image of resolution $s\mathrm{H} \times s\mathrm{W}$ is sampling a grid of $s^{2}$ rays in an original pixel. Obviously, not only the sampled ray directions were never seen during training, but the pixel queried corresponds to a smaller region in the 3D space. Regarding this issue, we propose a supersampling strategy that tackles the problem of rendering SR images for NeRF. The intuition of supersampling is explained as follows and illustrated in \figref{fig:super-sampling}.

We start from the image formation process. The pixel values are mapped from scene irradiance through a \textit{camera response function} (CRF). For simplification, we assume a pinhole camera model as in NeRF and consider ISO gain, shutter speed as implicit factors. Let $\mathcal{R}(\mathbf{p})$ denotes the set of all possible ray directions for pixel $\mathbf{p}$ from a training image, then:

\begin{equation}
    \mathcal{C}(\mathbf{p}) = f(E_{\mathcal{R}(\mathbf{p})})
\end{equation}
where $\mathcal{C}(\mathbf{p})$ indicates the color of $\mathbf{p}$, $f$ is CRF, $E$ is the incident irradiance over the area covered by $\mathbf{p}$, which is the integration of radiance over all incoming rays in $\mathcal{\mathbf{p}}$. Although ideally the training ray directions should be sampled from $\mathcal{R}(\mathbf{p})$, it is both computational expensive and challenging for the network to fit this huge amount of data. Therefore, in our work, to super-resolve images at the scale of $s$, we first evenly split a pixel from training set into a $s \times s$ grid sub-pixels $\mathcal{S}(\mathbf{p})$. As in NeRF, we do not model CRF and output the color of each sub-pixel using a multi-layer perceptron directly. During training stage, ray directions for a pixel $\mathbf{p}$ will be sampled from the sub-pixels instead, denoted as $\mathcal{R}^{\prime}(\mathbf{p}) = \{\mathbf{r}_{\mathbf{j}}\:|\: \mathbf{j} \in \mathcal{S}(\mathbf{p}) \} \subset \mathcal{R}(\mathbf{p})$. At inference stage, an $s\mathrm{H} \times s\mathrm{W}$ image can be directly obtained by directly rendering and organizing the sub-pixels, erasing the sampling gap between the training and testing phase.

Another concern is how to perform supervision with only ground truth images at dimension $\mathrm{H} \times \mathrm{W}$. Similar to the blind-SR problem, the degradation process from $s\mathrm{H} \times s\mathrm{W}$ is unknown and may be affected by many factors. Inspired by the graphics pipeline, we tackle this issue by compute the radiance for sub-pixels in $\mathcal{R}^{\prime}(p)$ using Equation \ref{equ:render} and then average them to compare with the color of $\mathbf{p}$. Thus, Equation \ref{equ:mse} can be extended as:
\begin{equation}
    \mathcal{L}_{\mathrm{MSE}} = \sum_{\mathbf{p} \in \mathcal{P}}\Big\| \frac{1}{|\mathcal{R}^{\prime}(\mathbf{p})|}\sum_{\mathbf{r}^{\prime} \in \mathcal{R}^{\prime}(\mathbf{p})}\hat{\mathbf{C}}(\mathbf{r}^{\prime}) - \mathbf{C}(\mathbf{r}_{\mathbf{p}}) \Big\|_2^{2}
    \label{equ:l_mse}
\end{equation}
where $\mathcal{R}^{\prime}(\mathbf{p})$ is the sub-pixel grid for pixel $\mathbf{p}$, $|\mathcal{R}^{\prime}(\mathbf{p})|$ is the number of sub-pixels in $\mathcal{R}^{\prime}(\mathbf{p})$, $\mathbf{r}^{\prime}$ is the ray direction for a single sub-pixel, $\hat{\mathbf{C}}(\mathbf{r}^{\prime})$ is the color of a sub-pixel predicted by the network. On the other hand, the LR images can be seen as downsampled from HR ones by averaging pixel color in a grid (We call it ``average'' kernel). This aborts any complex assumptions on the downsampling operation and make our method robust for various situations.

To summarize, supersampling extends original NeRF in two aspects: first it samples ray directions from $s \times s$ grid sub-pixels for pixel $\mathbf{p}$ instead of a single ray direction; second, it averages the color of the sub-pixels for supervision. In computer graphics, supersampling and averaging is often used in the rendering process to handle the problem of aliasing. In our work, we show that it fully exploits the cross-view consistency introduced by NeRF to a sub-pixel level, \ie a position can be corresponded through multiple viewpoints. While NeRF only shoots one ray for each pixel and optimizes points along that ray, supersampling constraints more positions in the 3D space and better utilize the multi-view information in input images. In other words, supersampling directly optimizes a denser radiance field at training time.

\begin{figure*}[htbp]
    \centering
    \includegraphics[width=1.0\linewidth]{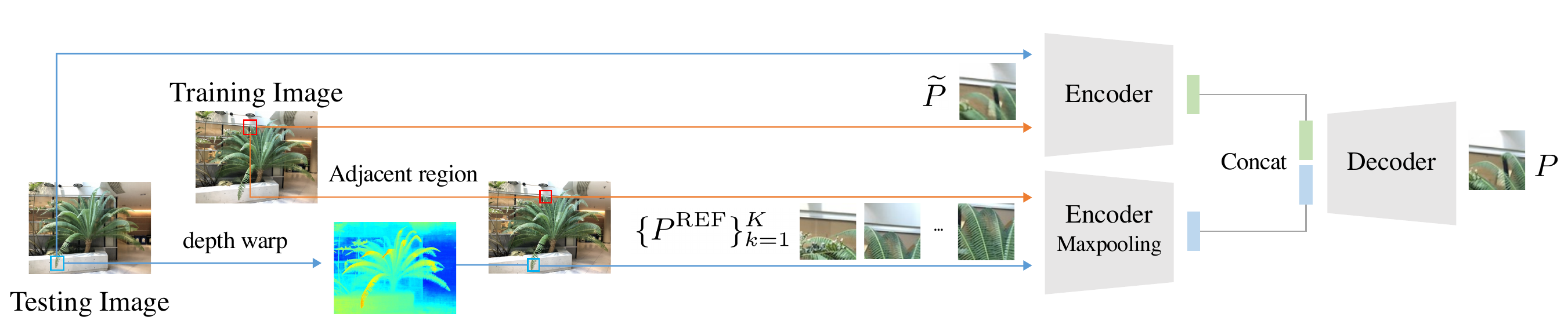}
    \caption{Our refinement module encodes synthesized patches $\widetilde{P}$ from images produced by supersampling and reference patches $\{P^{\mathrm{REF}}\}_{k=1}^{K}$ from $\mathcal{I}_{\mathrm{REF}}$. The encoded features of $\mathcal{I}_{\mathrm{REF}}$ are maxpooled and concatenated with that of $\widetilde{P}$, which is then decoded to generate the refined patch. In the training phase, $\widetilde{P}$ is sampled from synthesized SR image at the camera pose of $\mathcal{I}_{\mathrm{REF}}$ and $\{P^{\mathrm{REF}}\}_{k=1}^{K}$ is sampled at adjacent regions. When testing, $\{P^{\mathrm{REF}}\}_{k=1}^{K}$ is obtained via depth warping. (The input and output patches are zoomed for better illustration, zoom in to see the details on leaves after refinement)}
    \label{fig:refine}
\end{figure*}

\subsection{Patch-Based Refinement}
\label{subsec:refine}
With supersampling, the synthesized image achieves much better visual quality than vanilla NeRF. However, when the images for a scene do not have enough sub-pixel correspondence, the results of supersampling cannot find enough details for high-resolution synthesis. Also, often there are limited high-resolution images from which HR content are available for further improving the results.

Here, we present a patch-based refinement network to recover high-frequency details that works even in the \textit{extreme} case, \ie only one HR reference is available, as shown in \figref{fig:refine}. Our system is though not limited to one HR reference and can be easily extended to multiple HR settings. The core design consideration focuses on how to ``blend'' details on the reference image $\mathcal{I}_{\mathrm{REF}}$ into NeRF synthesized SR images that already captured the overall structure. We adopt a patch-by-patch refine strategy that turns an SR patch $\widetilde{P}$ into the refined patch $P$. Other than $\widetilde{P}$, the input should also include an HR patch from $\mathcal{I}_{\mathrm{REF}}$ that reveals how the objects or textures in $\widetilde{P}$ presents in high-resolution. However, due to occlusion and inaccuracy of depth estimation, multiple HR patches are required to cover the region in $\widetilde{P}$ and we use $K$ patches $\{P^{\mathrm{REF}}\}_{k=1}^{K}$  for reference. Also, patches in $\{P^{\mathrm{REF}}\}_{k=1}^{K}$ cover larger regions than $\widetilde{P}$ and contain less relevant information. The refinement stage aims at local detail enhancement and well preserve the view consistent structure from super-sampling with the spatial information of depth predictions.

We use a U-Net based convolutional architecture for the refinement network, which has demonstrated its efficacy in several existing novel view synthesis methods \cite{choi2019extreme, riegler2021stable, riegler2020free}. In earlier attempts, we model the refinement procedure as an image-to-image translation \cite{isola2017image} and find channel-wise stack $\widetilde{P}$ and $\{P^{\mathrm{REF}}\}_{k=1}^{K}$ were unable to fit the training set perfectly. Therefore, inspired by \cite{choi2019extreme, riegler2020free}, we instead encode each patch respectively with an encoder consisting of seven convolutional layers. The decoder of the network takes as input the nearest-neighbor upsampled features from previous layers concatenated with both the encoded features of $\widetilde{P}$ and maxpooled features of $\{P^{\mathrm{REF}}\}_{k=1}^{K}$ at the same spatial resolution. All convolutional layers are followed by a ReLU activation. 

\topic{Training} 
The training of the refinement network requires SR and HR patch pairs, which are only available at the camera pose of $\mathcal{I}_{\mathrm{REF}}$. Therefore, $\widetilde{P}$ is randomly sampled from the SR image and $P$ is the patch on $\mathcal{I}_{\mathrm{REF}}$ at the same location.  We perform perspective transformations to $\widetilde{P}$ and $P$ as during testing, the input patches are mostly from different camera poses. Moreover, to account for the inaccuracy of reference patches at testing time, we sample $\{P^{\mathrm{REF}}\}_{k=1}^{K}$ within a fixed window around $P$. In order to preserve the spatial structure of $\widetilde{P}$ while improving its quality, our objective function combines reconstruction loss $\mathcal{L}_{rec}$ and perceptual loss $\mathcal{L}_{per}$, where 
\begin{equation}
\mathcal{L_\mathrm{refine}} = \mathcal{L}_{rec} + \mathcal{L}_{per} = ||\widetilde{P} - P ||_1 + \Sigma_{l}\lambda_{l}||\phi_{l}(\widetilde{P}) - \phi_{l}(P) ||_1
\end{equation}
$\phi_{l}$ is a set of layers in a pretrained VGG-19 and $\lambda{l}$ is the reciprocal of the number of neurons in layer $l$. Note that we adopt $\mathnormal{l}_1$-norm instead of MSE in $L_{rec}$ because it is already minimized in supersampling and $\mathnormal{l}_1$-norm will sharpen the results. 

\topic{Testing} At inference time, given a patch $\widetilde{P}$ on synthesized image $\mathcal{I}_n$, we can find a high-resolution reference patch on reference image $\mathcal{I}_{\mathrm{REF}}$ for each pixel on $\widetilde{P}$:
\begin{equation}
    P_{i,j}^{\mathrm{REF}} = K_{\mathrm{REF}}T(K_{n}^{-1}d_{i,j}\widetilde{P}_{i,j})
    \label{equ:warp}
\end{equation}
where $i,j$ denotes a location on patch $\widetilde{P}$, $d$ is the estimated depth, $T$ is the transformation between camera extrinsic matrices from $\mathcal{I}_n$ to $\mathcal{I}_{\mathrm{REF}}$, and $K_{\mathrm{REF}}$ and $K_{n}$ refer to the camera intrinsic matrices of $\mathcal{I}_{\mathrm{REF}}$ and $\mathcal{I}_n$. Therefore, \eqnref{equ:warp} computes the 3D world coordinate of $i,j$ based on $d_{i,j}$ and camera parameters, then backproject it to a pixel on $\mathcal{I}_{\mathrm{REF}}$ and extract the corresponding patch at that location (points fall out of $\mathcal{I}_{\mathrm{REF}}$ are discarded). In summary, to obtain the refined $P$, we first sample $K$ patches from $\{P_{i,j}^{\mathrm{REF}}\}$ to construct the set $\{P^{\mathrm{REF}}\}_{k=1}^{K}$ and then input them together with $\widetilde{P}$ into the network. 
More details of the refinement network can be found in the \href{\suppurl}{supplementary material}.

The training of NeRF requires correspondences of input images in the 3D space. As long as the HR reference falls in the camera frustum of input images, it can be easily wrapped to other views and bring enough details. Therefore, our refinement network is well-suited for any NeRF compatible dataset.


\section{Experiments}
\label{sec:experiments}

\begin{table*}[htbp]
\centering
\resizebox{\textwidth}{!}{%
\begin{tabular}{l|ccc|ccc|ccc|ccc}
& \multicolumn{3}{c|}{Blender$\times 2$ ($100 \times 100$)} & \multicolumn{3}{c|}{Blender$\times 4$ ($100 \times 100$)} & \multicolumn{3}{c}{Blender$\times 2$ ($200 \times 200$)} & \multicolumn{3}{c}{Blender$\times 4$ ($200 \times 200$)} \\
Method & PSNR$\uparrow$ & SSIM$\uparrow$ & LPIPS$\downarrow$ & PSNR$\uparrow$ & SSIM$\uparrow$ & LPIPS$\downarrow$ & PSNR$\uparrow$ & SSIM$\uparrow$ & LPIPS$\downarrow$ & PSNR$\uparrow$ & SSIM$\uparrow$ & LPIPS$\downarrow$ \\
\hline
NeRF~\cite{mildenhall2020nerf} & $\underline{27.54}$ & $\underline{0.921}$ & $0.100$ & $\underline{25.56}$ & $0.881$ & $0.170$ & $\underline{29.16}$ & $\underline{0.935}$ & $0.077$ & $\underline{27.47}$ & $0.910$ & $0.128$  \\
NeRF-Bi & $26.42$ & $0.909$ & $0.151$ & $24.74$ & $0.868$ & $0.244$ & $28.10$ & $0.926$ & $0.109$ & $26.67$ & $0.900$ & $0.175$  \\
NeRF-Liif & $27.07$ & $0.919$ & $\underline{0.067}$ & $25.36$ & $\underline{0.885}$ & $0.125$ & $28.81$ & $0.934$ & $\underline{0.058}$ & $27.34$ & $\underline{0.912}$ & $0.096$  \\
NeRF-Swin & $26.34$ & $0.913$ & $0.075$ & $24.85$ & $0.881$ & $\underline{0.108}$ & $28.03$ & $0.926$ & $\underline{0.058}$ & $26.78$ & $0.906$ & $\underline{0.086}$  \\
Ours-SS & $\boldsymbol{29.77}$ & $\boldsymbol{0.946}$ & $\boldsymbol{0.045}$ & $\boldsymbol{28.07}$ & $\boldsymbol{0.921}$ & $\boldsymbol{0.071}$ & $\boldsymbol{31.00}$ & $\boldsymbol{0.952}$ & $\boldsymbol{0.038}$ & $\boldsymbol{28.46}$ & $\boldsymbol{0.921}$ & $\boldsymbol{0.076}$ 
\end{tabular} 
} 
\caption{Quality metrics for novel view synthesis on blender dataset. We report PSNR/SSIM/LPIPS for scale factors $\times2$ and $\times4$ on two input resolutions ($100 \times 100$ and $200 \times 200$) respectively. 
}
\label{table:blender-results}  
\end{table*}
\setlength{\tabcolsep}{1.4pt}

In this section, we provide both quantitative and qualitative comparisons to demonstrate the advantages of the proposed \sysname{}. We first show results and analysis of super-sampling, and then demonstrate how the refinement network adds more details to it. Our result only with super-sampling is denoted as Ours-SS and our result after patch-based refinement is denoted as Ours-Refine.

\subsection{Dataset and Metrics}
To evaluate our methods, we train and test our model on the following datasets. We evaluate the quality of view synthesis with respect to ground truth from the same pose using three metrics: Peak Signal-to-Noise Ratio (PSNR) and Structural Similarity Index Measure (SSIM) \cite{wang2003multiscale} and LPIPS\cite{zhang2018unreasonable}.

\topic{Blender Dataset} The Realistic Synthetic $360^{\circ}$ of \cite{mildenhall2019local} (known as Blender dataset) contains 8 detailed synthetic objects with 100 images taken from virtual cameras arranged on a hemisphere pointed inward. As in NeRF\cite{mildenhall2020nerf}, for each scene we input 100 views for training and hold out 200 images for testing.

\topic{LLFF Dataset} LLFF dataset\cite{mildenhall2019local, mildenhall2020nerf} consists of 8 real-world scenes that contain mainly forward-facing images. We train on all the images and report the average metrics on the whole set.

\subsection{Training Details}
In super-sampling, we implement all experiments on top of NeRF~\cite{mildenhall2020nerf} using PyTorch. As we train on different image resolutions independently, for fair comparison we train blender dataset and LLFF dataset for respectively 20 epochs and 30 epochs, where each epoch contains an iteration of the whole training set. We choose Adam as the optimizer (with hyperparameters $\beta_1 = 0.9$, $\beta_2 = 0.999$) with batch size set to 2048 (2048 rays a batch for all experimented scales) and learning rate decayed exponentially from $5\cdot 10^{-4}$ to $5 \cdot 10^{-6}$. Following NeRF, \sysname{} also uses a hierarchical sampling with the same size ``coarse'' and ``fine'' MLP. The number of coarse samples and fine samples are both set to 64.

\begin{figure*}[htbp]
    \centering
    \includegraphics[width=1.0\linewidth]{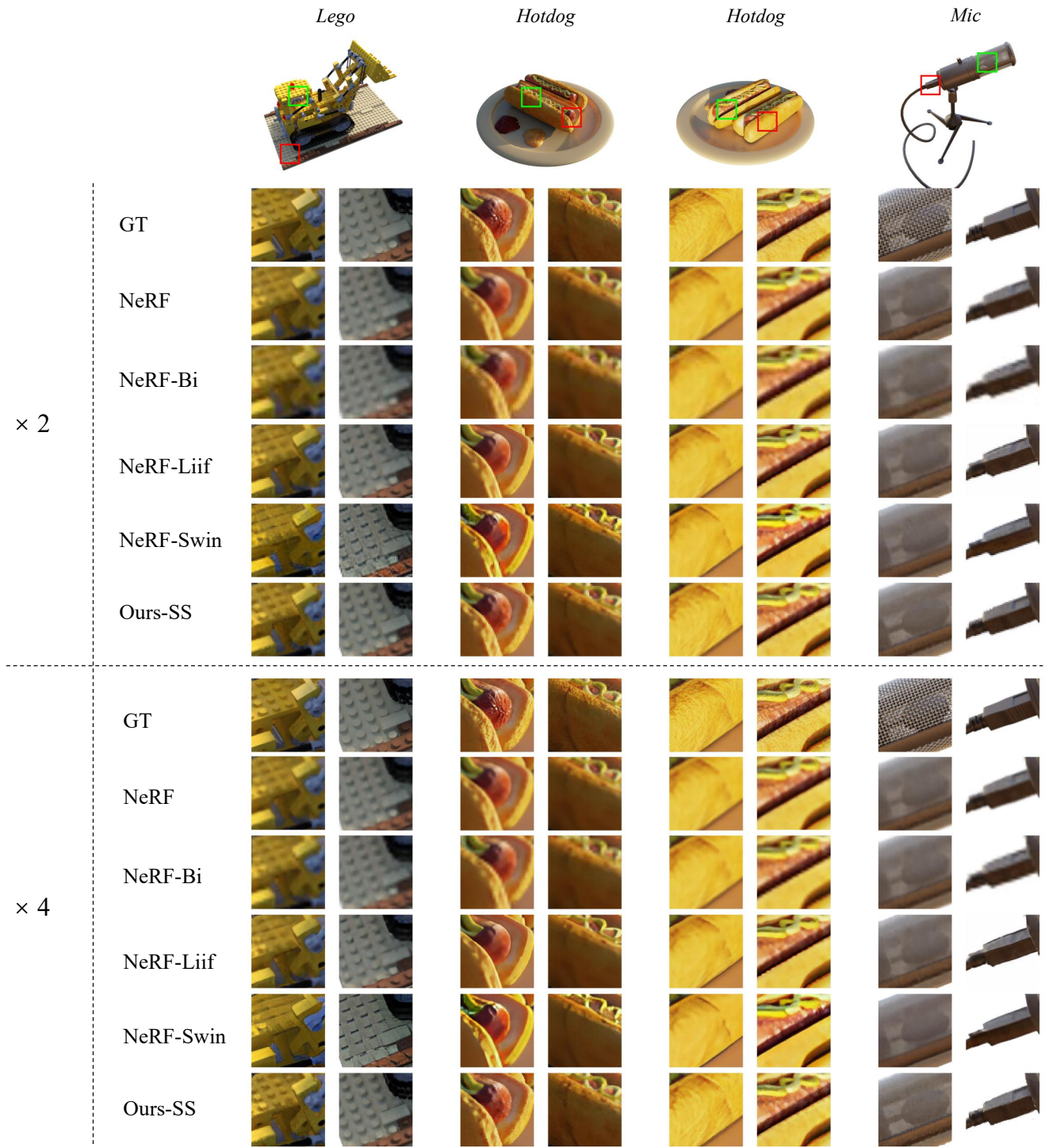}
    \caption{Qualitative comparison of blender dataset when the input images are $200 \times 200$ and upscale by 2 and 4. Note how \sysname{} recovers correct details through super-sampling even when inputting low-resolution images, such as \textit{Lego}'s gears, \textit{Hotdog}'s sausage and sauce, \textit{Mic}'s magnets and shiny brackets. Note \sysname{} is able to synthesize consistently over different viewpoints, here we provide two for \textit{Hotdog}, videos can be found on our \href{\weburl}{website}. Please zoom in for a better inspection of the results. }
    \label{fig:res-blender}
\end{figure*}

\newcommand{\resultsfigwidth}{1.0in}
\newcommand{\resultscropwidth}{0.85in}
\newcommand{\croporchid}[1]{
  \makecell{
  \includegraphics[trim={512px 244px 432px 448px}, clip, width=\resultscropwidth]{#1} \\
  \includegraphics[trim={93px 339px 801px 303px}, clip, width=\resultscropwidth]{#1} 
  }
}

\newcommand{\cropflower}[1]{
  \makecell{
  \includegraphics[trim={370px 310px 518px 326px}, clip, width=\resultscropwidth]{#1} \\
  \includegraphics[trim={550px 506px 338px 130px}, clip, width=\resultscropwidth]{#1} 
  }
}

\newcommand{\crophorns}[1]{
  \makecell{
  \includegraphics[trim={384px 280px 478px 330px}, clip, width=\resultscropwidth]{#1}   \\
  \includegraphics[trim={486px 256px 408px 386px}, clip, width=\resultscropwidth]{#1} 
  }
}

\begin{figure*}[htbp]
\centering
\scriptsize
\begin{tabular}{@{}c@{\,\,}c@{}c@{}c@{}c@{}c@{}c@{}c@{}}
\setlength{\tabcolsep}{10pt}
\makecell[c]{
\includegraphics[trim={0px 0px 0px 0px}, clip, width=\resultsfigwidth]{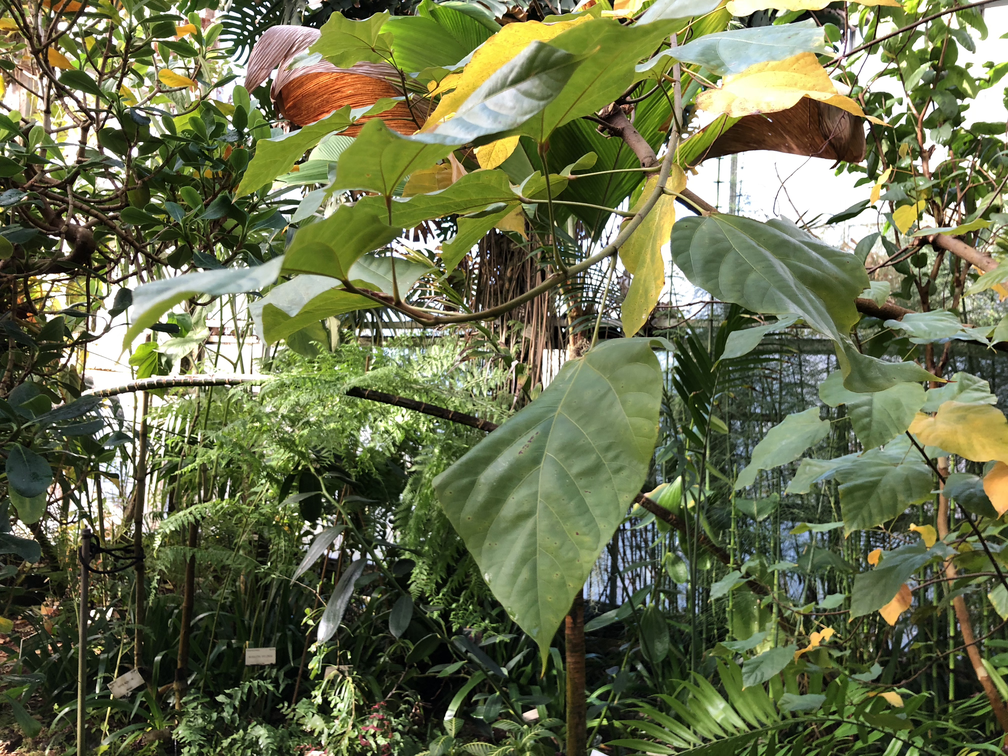}
\\
\textit{Leaves}
}
& 
\croporchid{results/leaves/gtx4.png} &
\croporchid{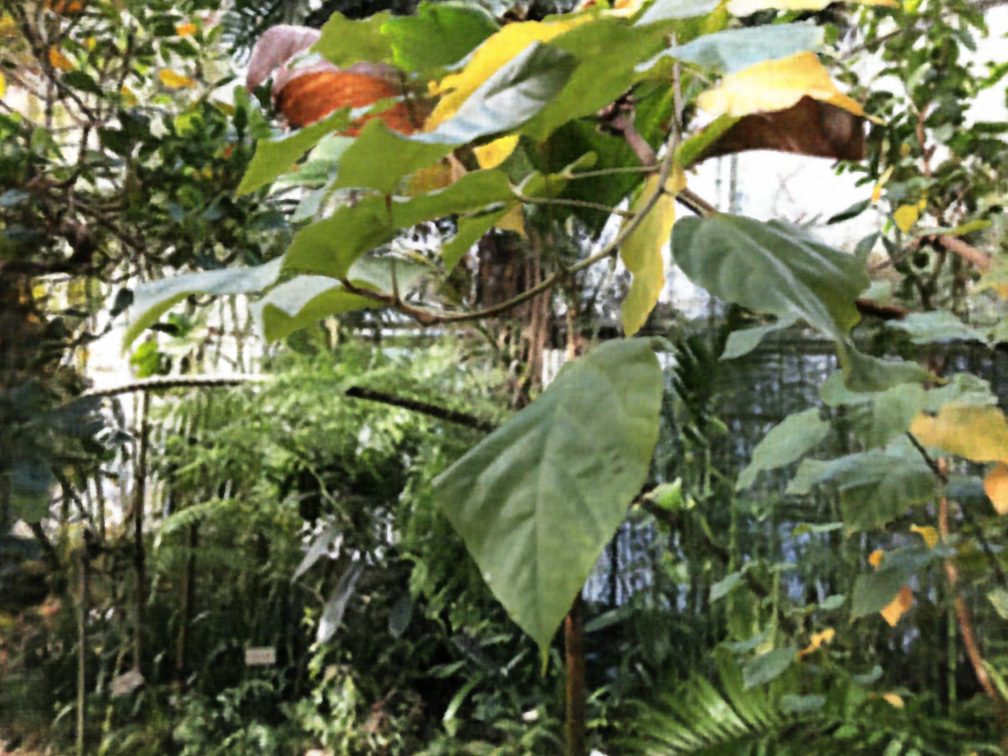} &
\croporchid{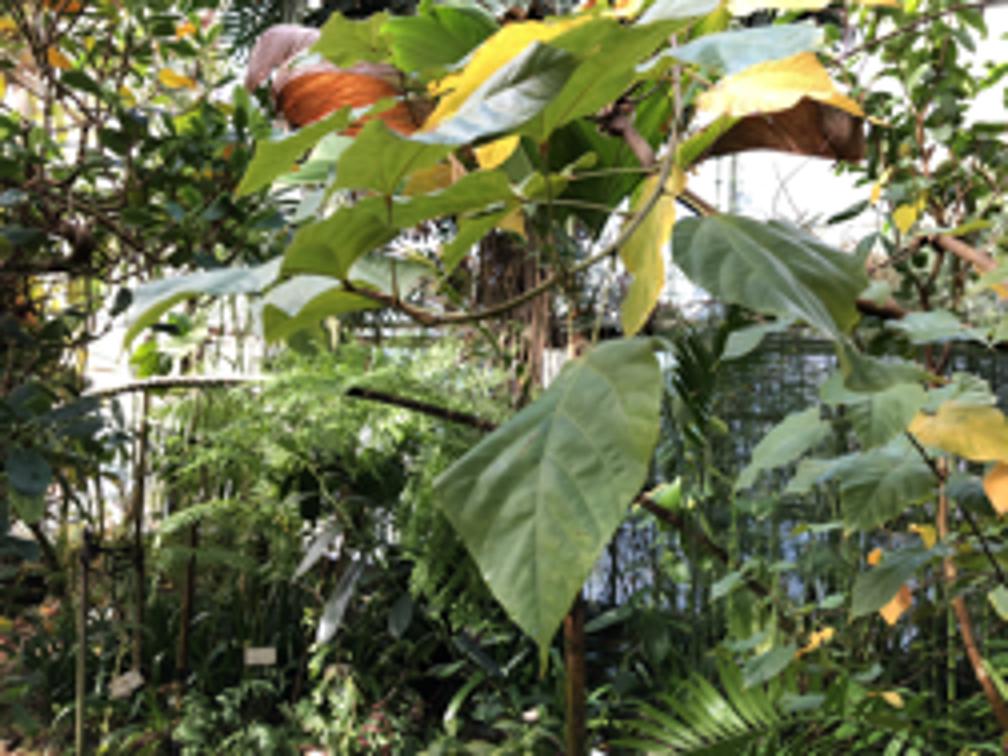} &
\croporchid{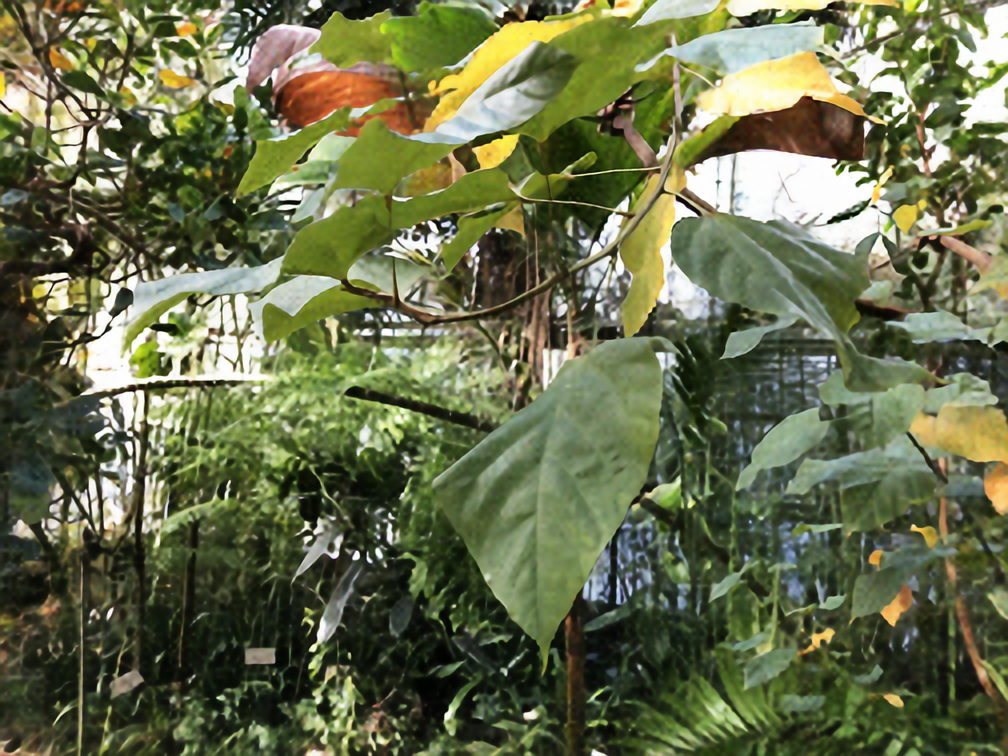} &
\croporchid{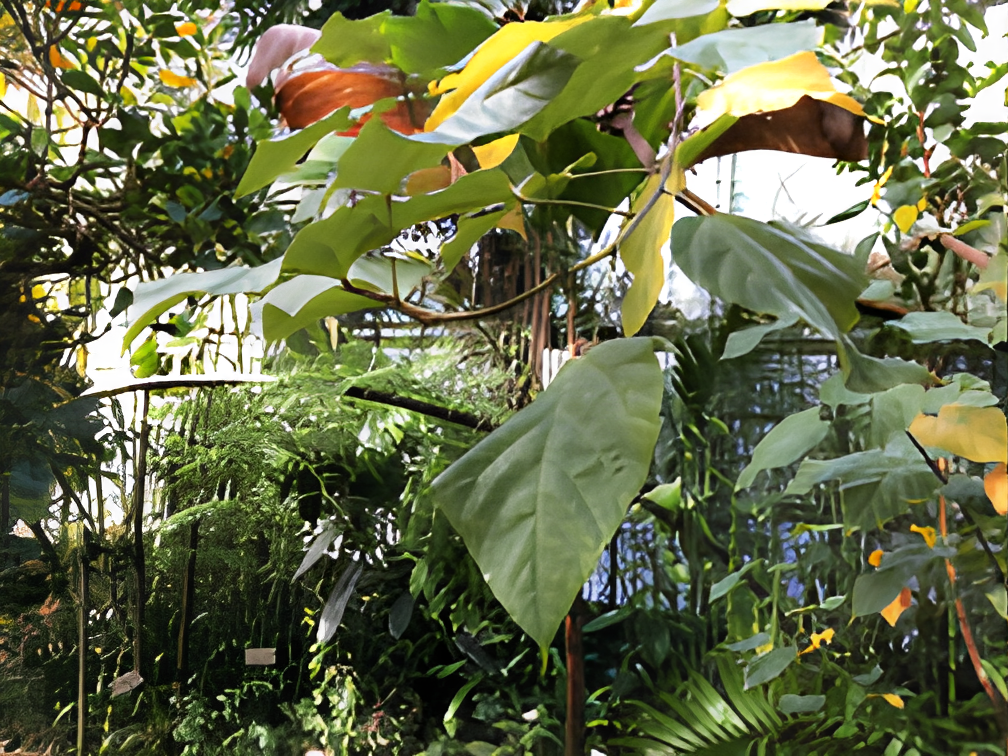} &
\croporchid{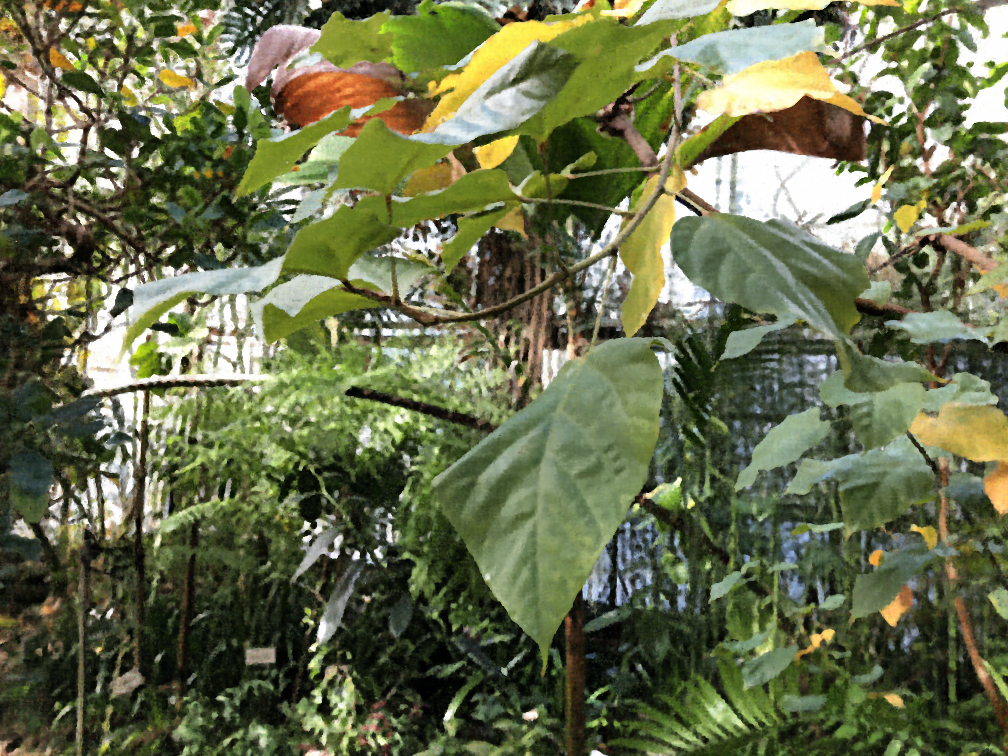} &
\croporchid{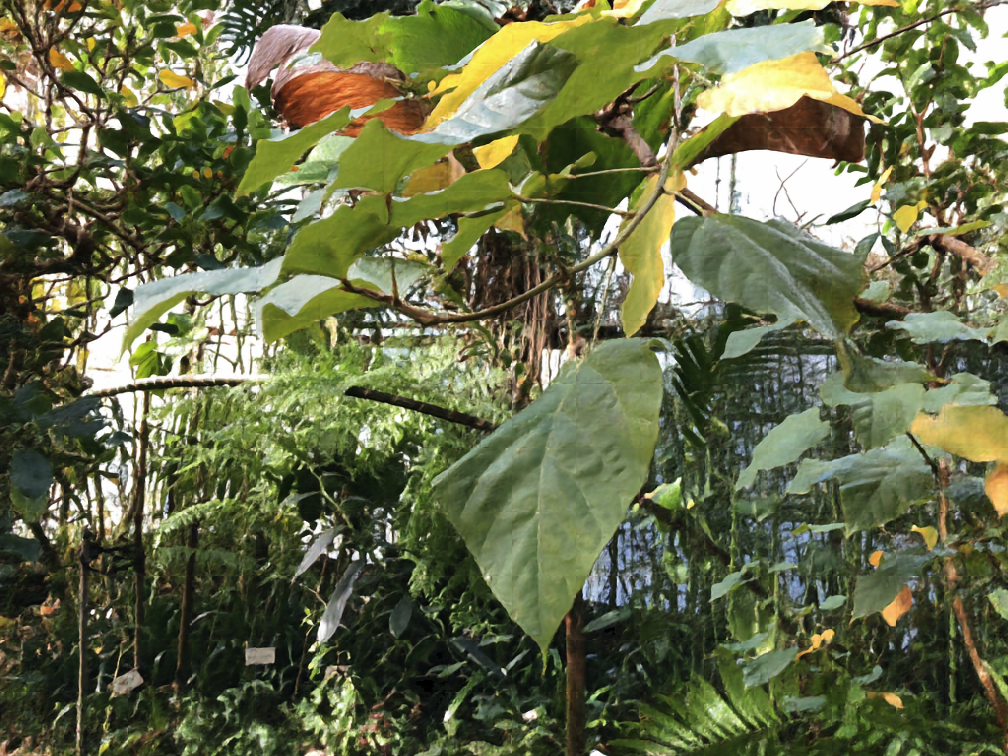}
\\
\makecell[c]{
\includegraphics[trim={0px 0px 0px 0px}, clip, width=\resultsfigwidth]{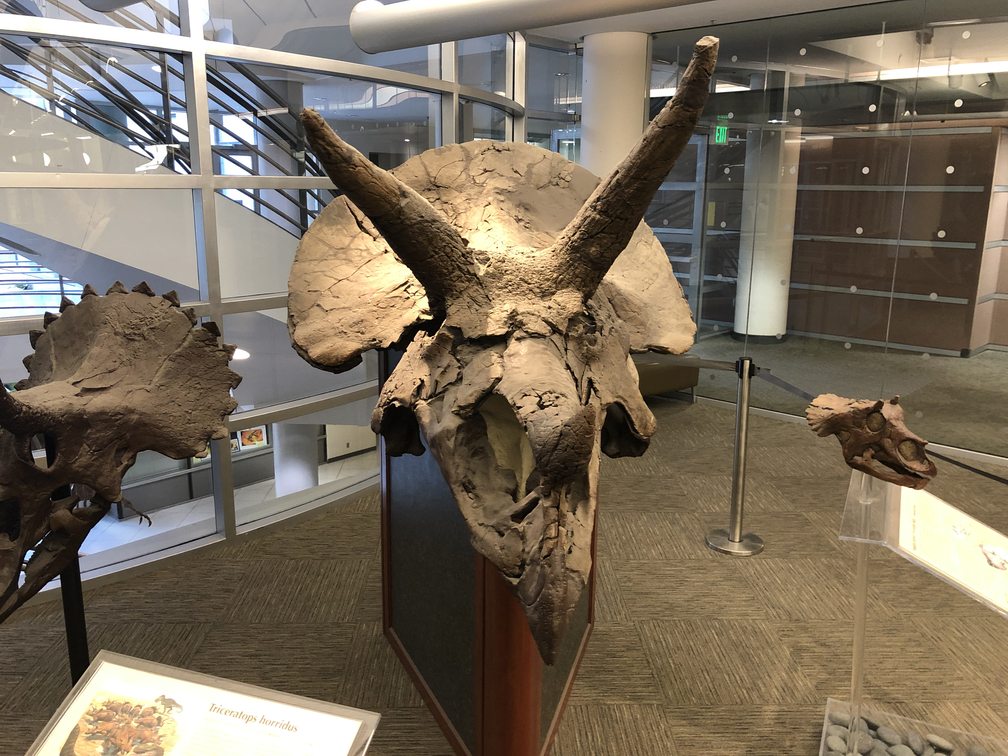}
\\
\textit{Horns}
}
& 
\crophorns{results/horns/gtx4.png} &
\crophorns{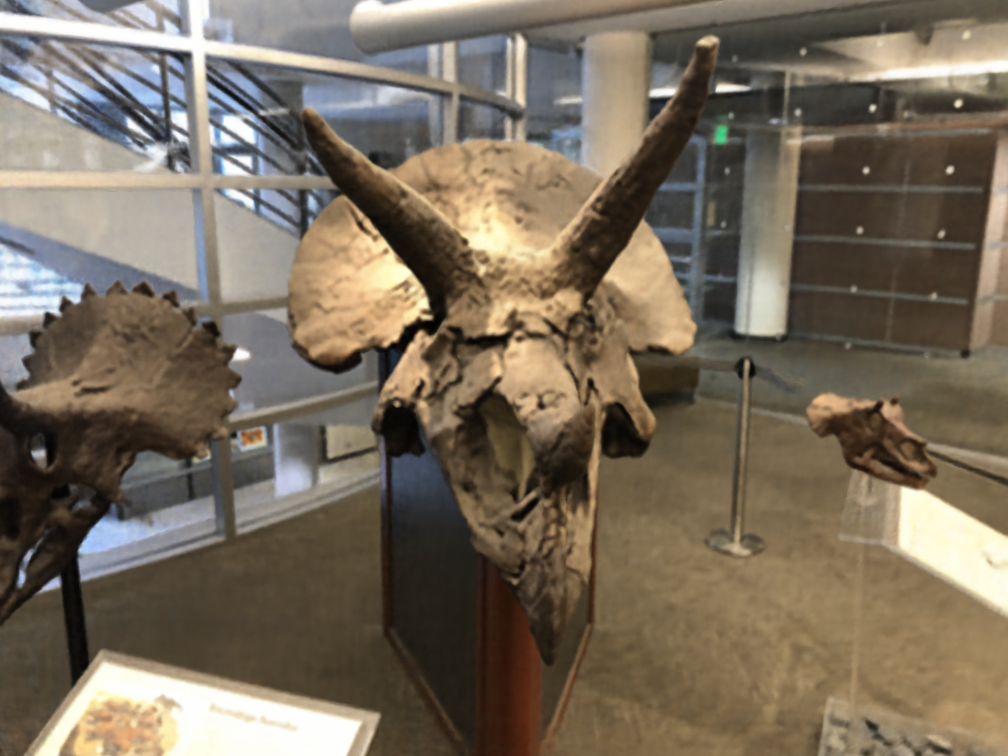} &
\crophorns{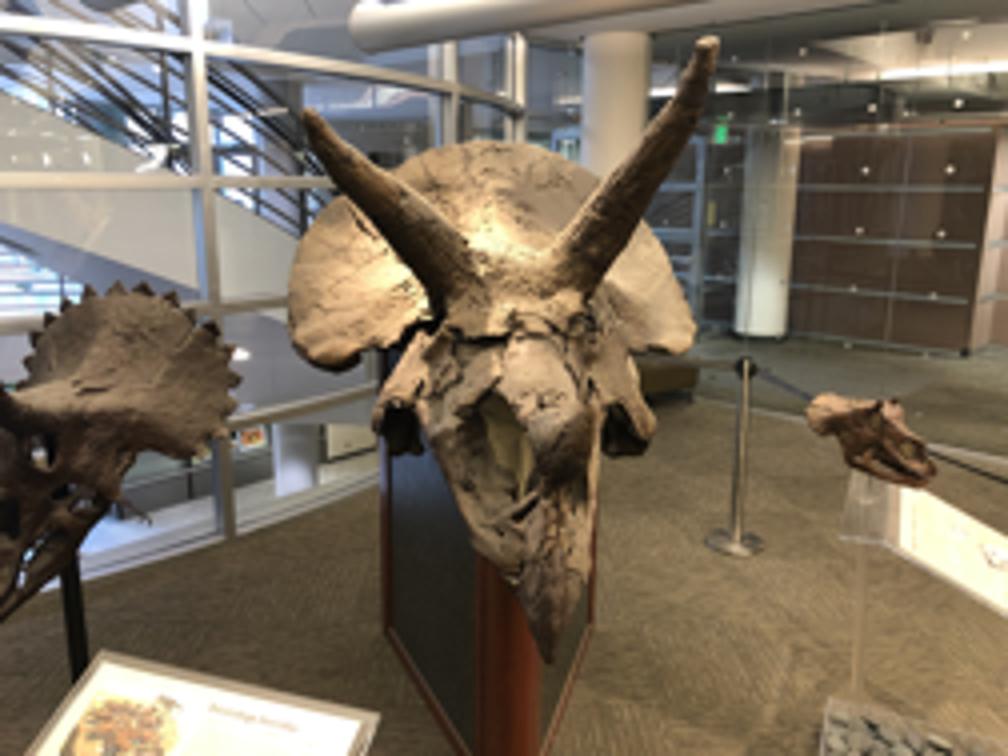} &
\crophorns{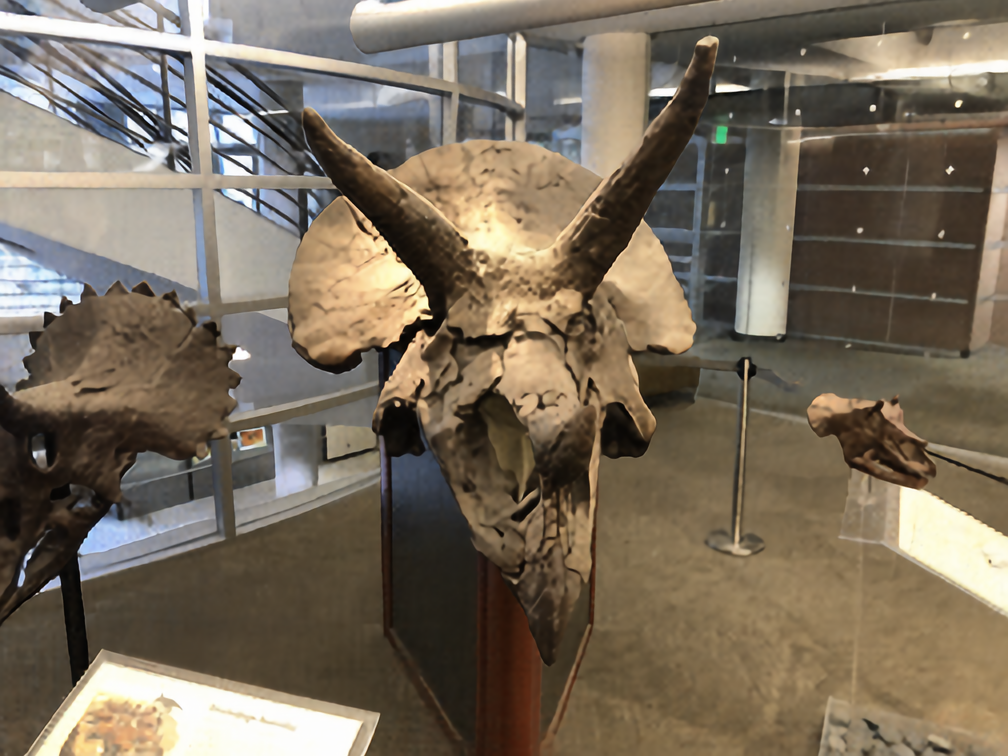} &
\crophorns{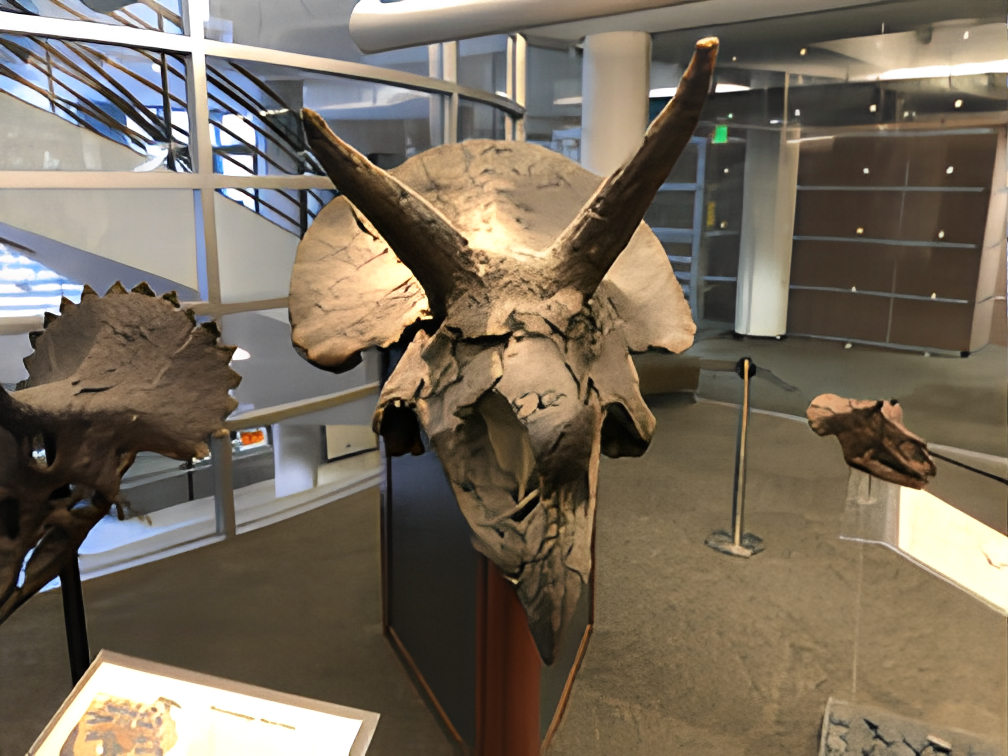} &
\crophorns{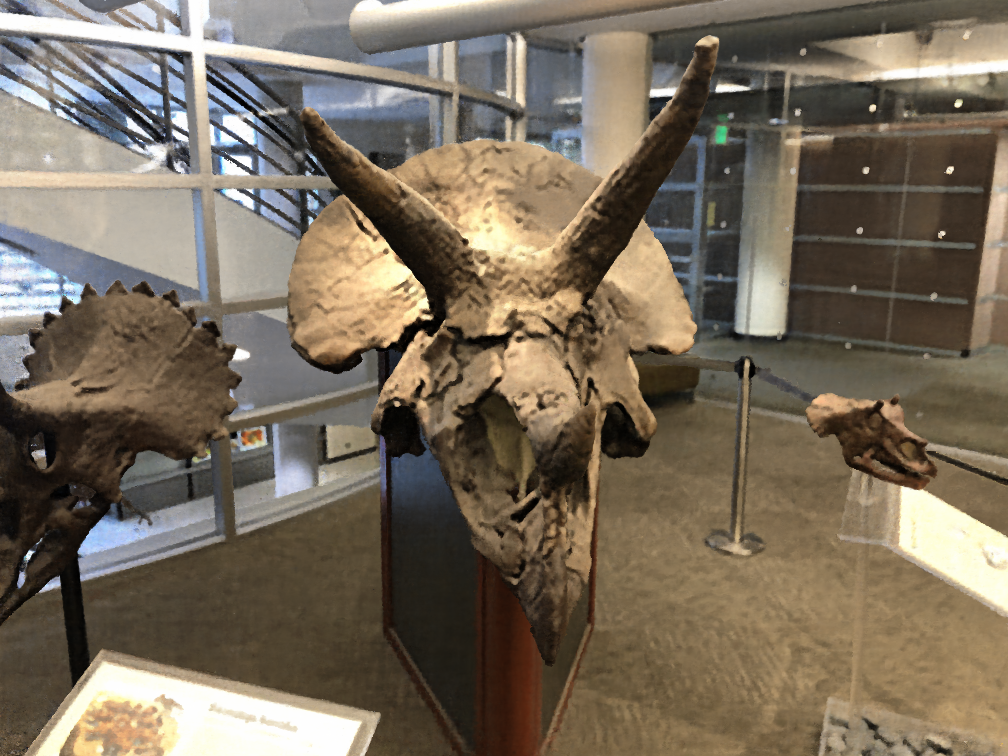} &
\crophorns{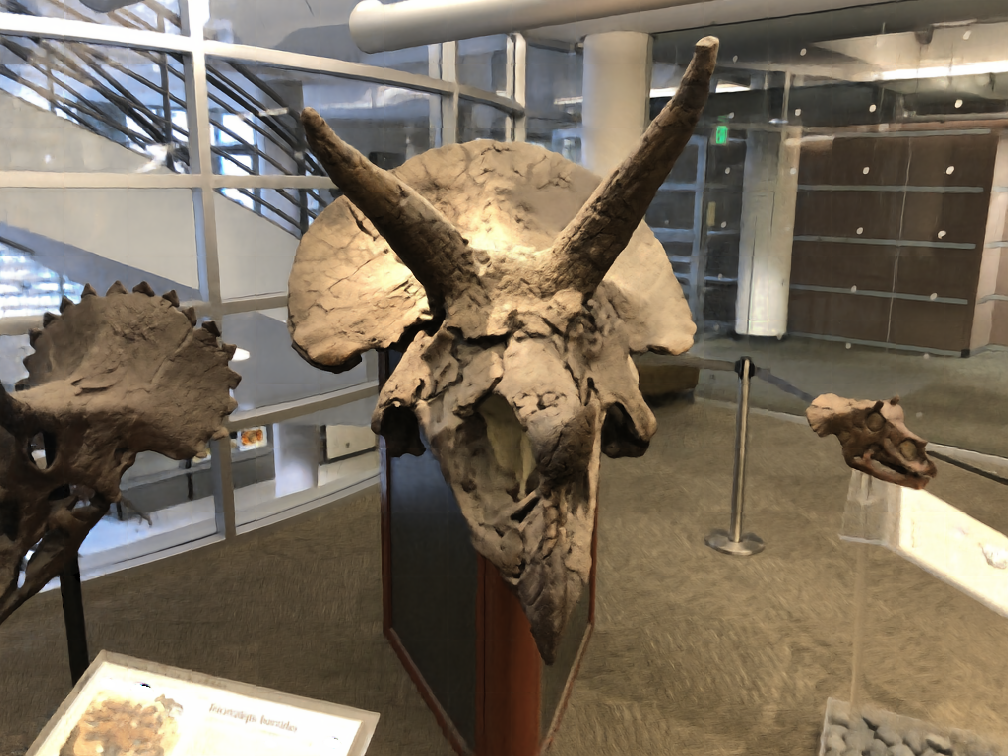}
\\
\makecell[c]{
\includegraphics[trim={0px 0px 0px 0px}, clip, width=\resultsfigwidth]{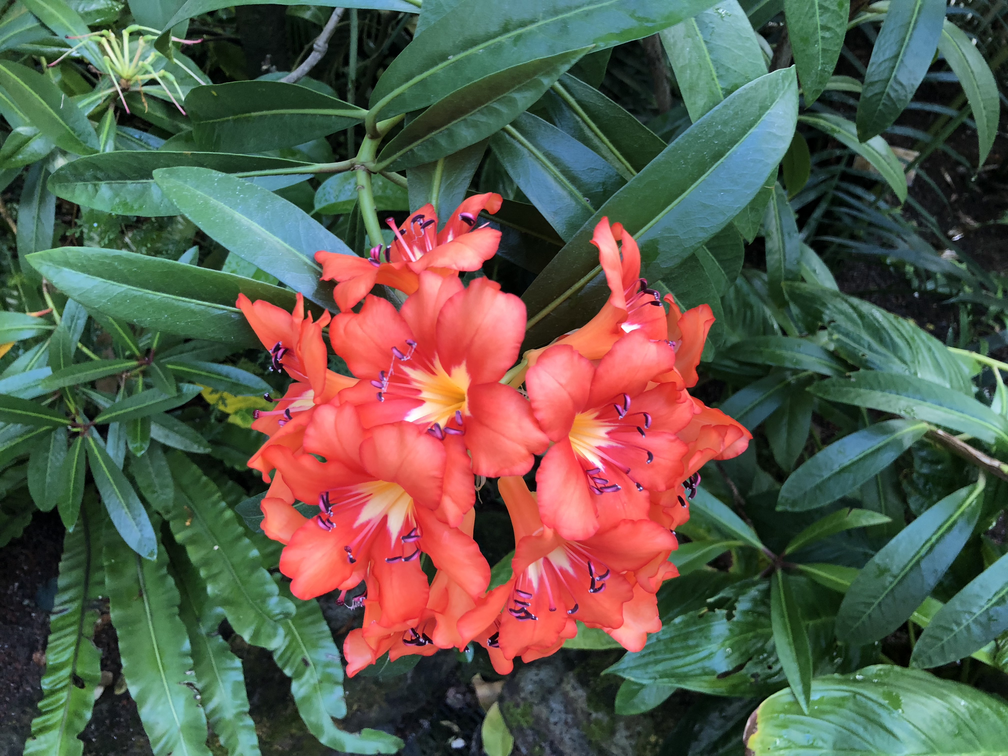}
\\
\textit{Flower}
}
& 
\cropflower{results/flower/gtx4.png} &
\cropflower{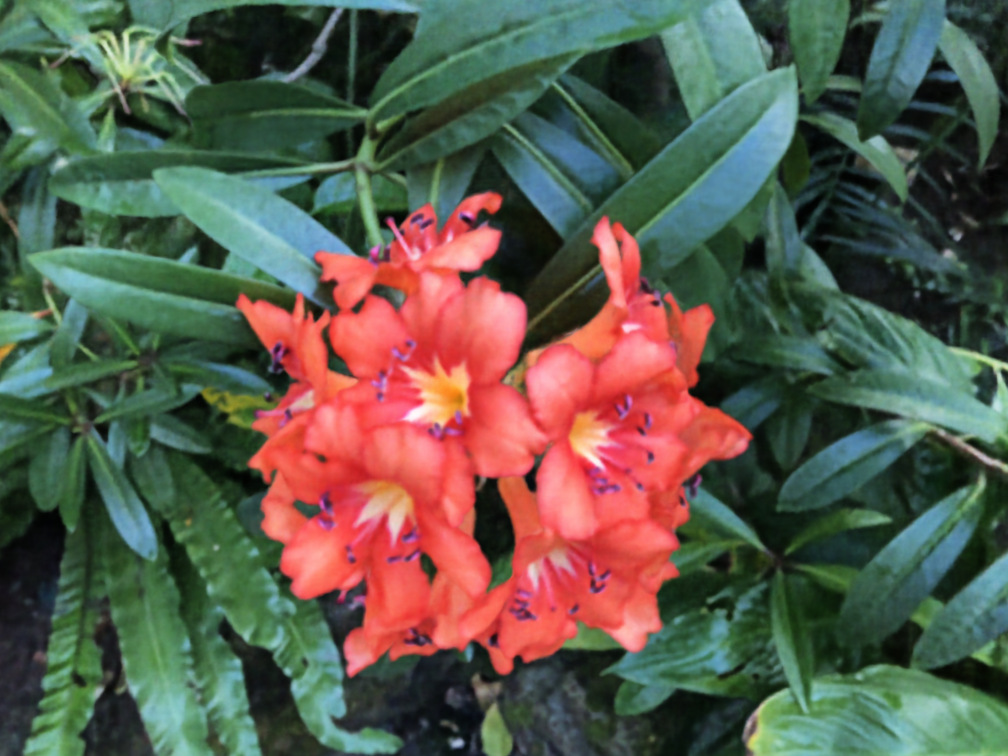} &
\cropflower{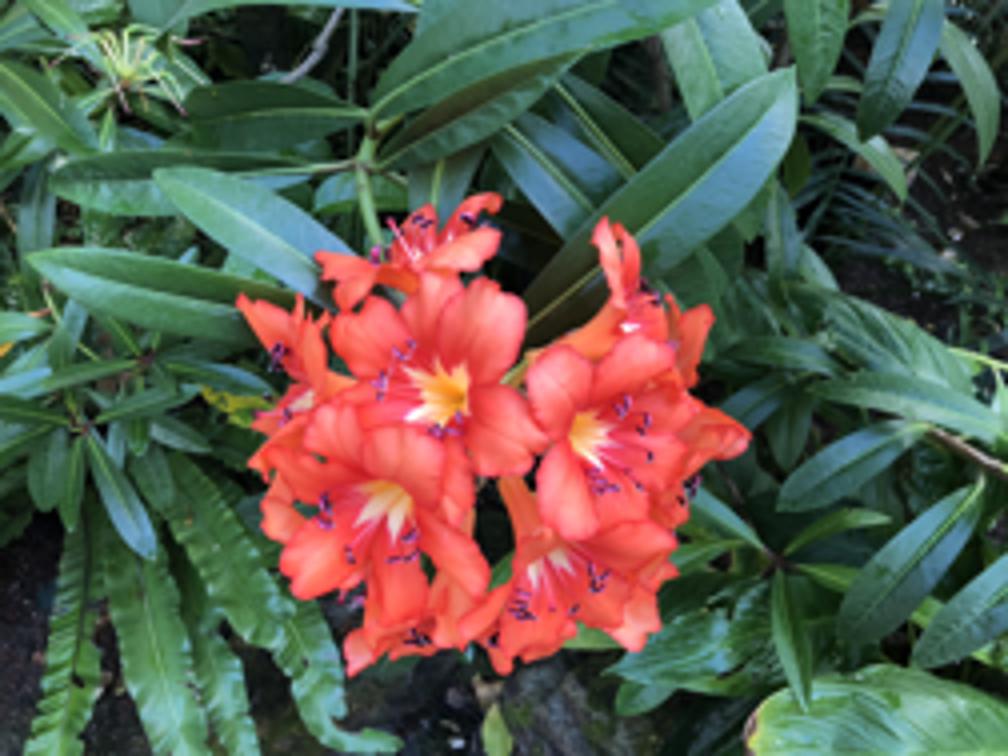} &
\cropflower{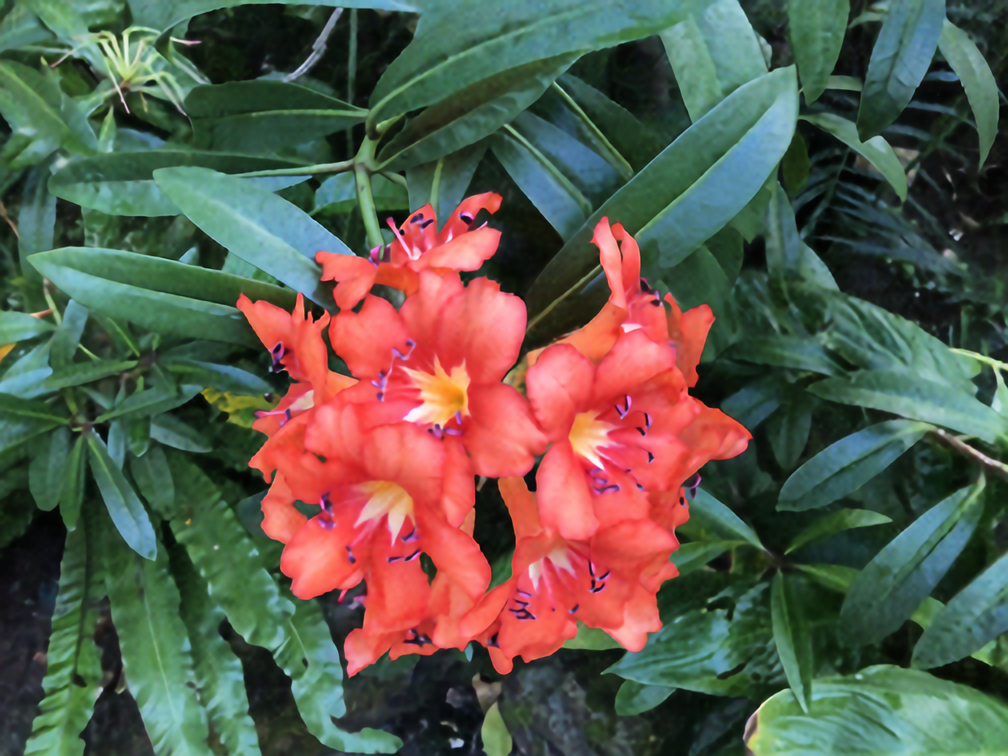} &
\cropflower{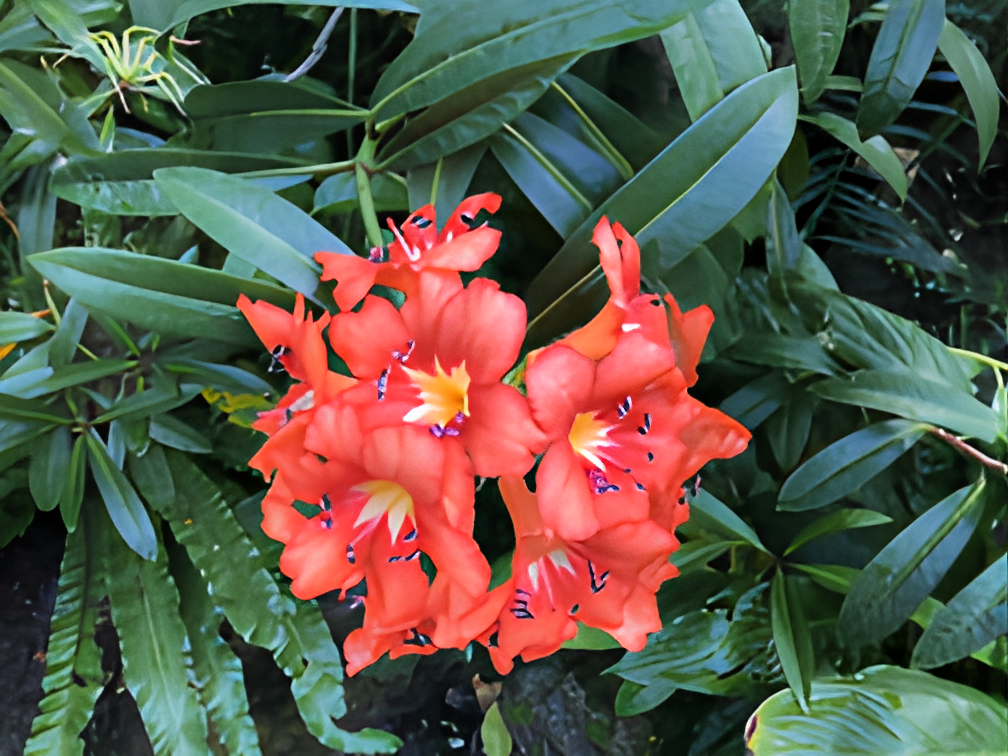} &
\cropflower{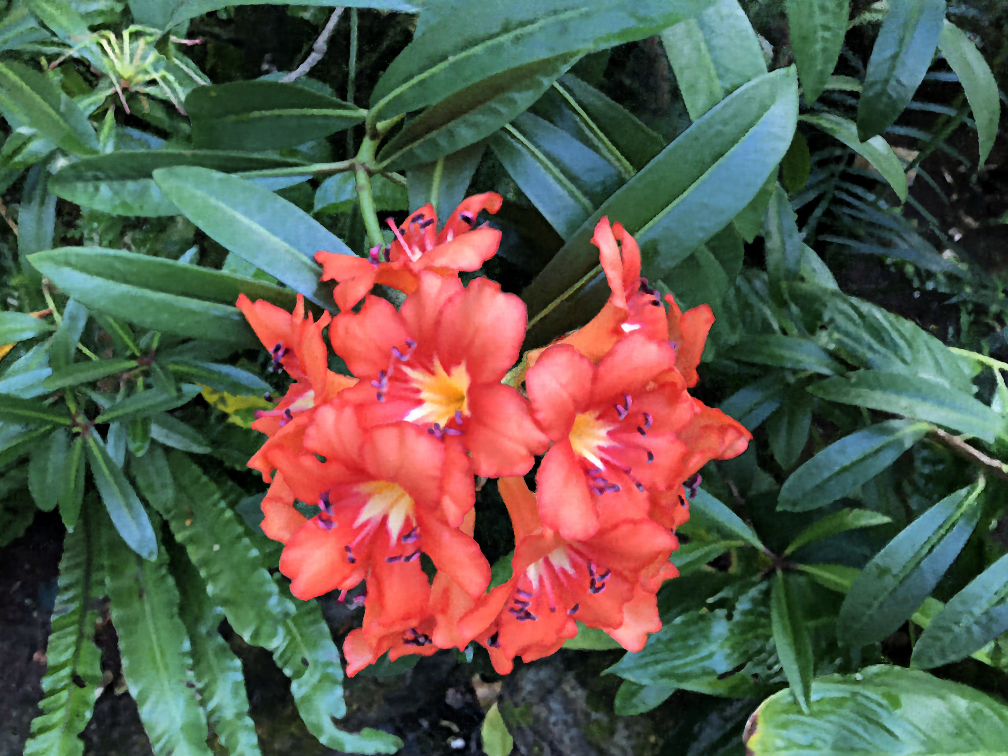} &
\cropflower{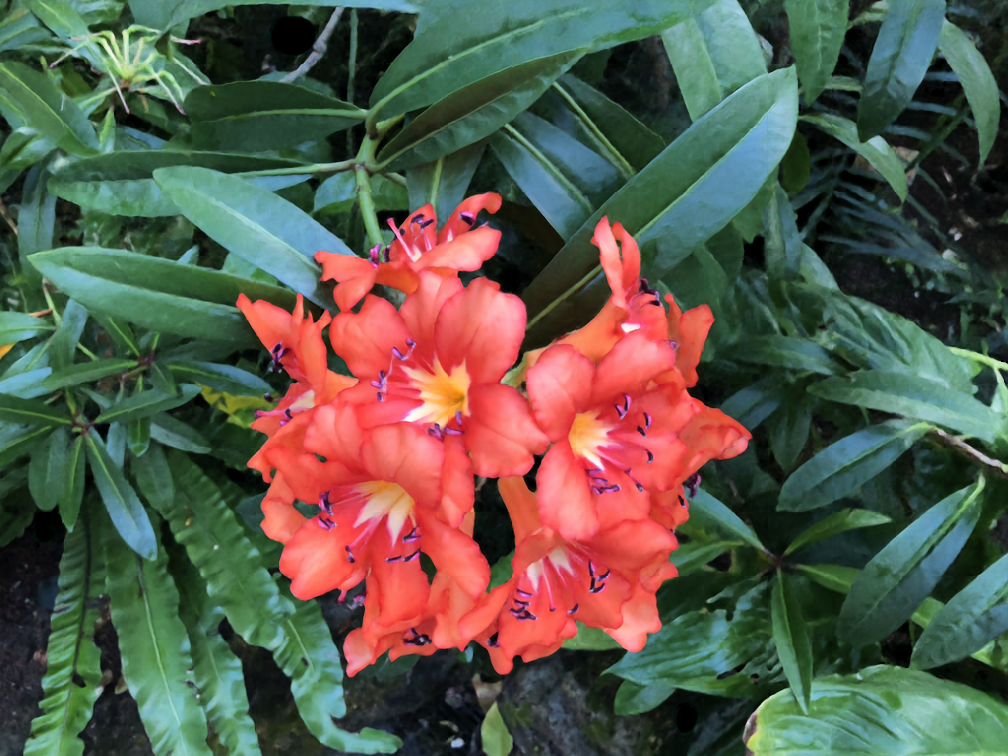}
\\
& Ground Truth & NeRF & NeRF-Bi & NeRF-Liif & NeRF-Swin & Ours-SS & Ours-Refine
\end{tabular}
\caption{Qualitative comparison on LLFF dataset at an upscale of 4 between bicubic, NeRF, Ours-SS and Ours-Refine. \sysname{} presents correct and clear texture in the leaves of \textit{Leaves} and \textit{Flowers} and fissures on \textit{Horns}' ears and noses, which can further be enhanced
using the refinement network. Please zoom in for better inspection of the results.}
\label{fig:res-llff}
\end{figure*}

\subsection{Comparisons}
Since there are no previous work that deals with super-resolving NeRF, we devise several reasonable baselines for comparisons, detailed as the following:

\topic{NeRF} Vanilla NeRF is already capable of synthesising images at any resolution due to its implicit formation. Therefore, we train NeRF on LR inputs using the same hyperparameters in our method and directly render HR images.

\topic{NeRF-Bi} aims to super-resolve a trained LR NeRF. We use the same trained model in the NeRF baseline, but render LR images directly and upsample them with the commonly used bicubic upsampling.

\topic{NeRF-Liif} Liif~\cite{chen2021learning} achieves state-of-the-art performance on continuous single image super-resolution. Similar to the NeRF-Bi baseline, we super-resolve LR images using pretrained liif model instead. Note that to the training process of liff requires LR-HR pairs, therefore it introduces external data priors.

\topic{NeRF-Swin} SwinIR~\cite{liang2021swinir} is the start-of-the-art method on single image super-resolution. Like NeRF-Bi and NeRF-Liif, NeRF-Swin performs super-resolution on a LR NeRF with the released SwinIR models under the ``Real-World Image Super-Resolution'' setting, which has a training set of more than 10k LR-HR pairs.


\subsection{Effectiveness of supersampling}
For blender dataset, we super-sample on two resolutions: $100 \times 100$ and $200 \times 200$, and test scales $\times 2$ and $\times 4$. For the LLFF dataset, the input resolution is $504 \times 378$ and we also upscale by $\times 2$ and $\times 4$. The downscaling of images in the dataset from original resolution to training resolution is done by the default Lanczos method in the Pillow package.

\figref{fig:res-blender} shows qualitative results for all methods on a subset of blender scenes. Renderings from NeRF-Bi exhibit correct global shapes but lack high-frequency details. Vanilla NeRF produces renderings that have more details than NeRF-Bi if the scene is already well-reconstructed at input resolution. However, it is still restricted by the information in the input image. NeRF-Liif can recover some details, but lacks enough texture. \sysname{} find sub-pixel level correspondence through supersampling, which means missing details in the input can be found from other views that lie in the neighboring region in 3D space. 

Quantitative results of blender dataset are summarized in \tabref{table:blender-results}. \sysname{} outperforms other baselines in all scenarios. NeRF-Liif or NeRF-Swin have the second best LPIPS, providing good visual quality but cannot even compete with NeRF in PSNR and SSIM. The reason is maybe the blender dataset is synthetic and has different domain than the dataset it is trained on, resulting false prediction (see NeRF-Swin on \textit{Lego} and \textit{Hotdog}).

The qualitative and quantitative results for LLFF dataset are demonstrated in \figref{fig:res-llff} and \tabref{table:llff-results} respectively. NeRF and NeRF-Bi suffers from blurry outputs. While NeRF-Liif and NeRF-Swin recovers some details and achieve satisfying visual quality (comparable LPIPS to Ours-SS) since they are trained on external datasets, they tend to be oversmooth and even predicts false color or geometry (See the leaves of \textit{Flower} in \figref{fig:res-llff}). \sysname{} fill in the details on the complex scenes and outperforms other baselines significantly. Therefore, we can conclude that learning-based 2D baselines struggle to perform faithfully super-resolution, especially in the multi-view case. 

In \secref{subsec:ss}, we mentioned that the supervision is performed by comparing the average color of sub-pixels due to the unknown nature of the degradation process (We call it ``average kernel''). However, in our experiments, the degradation kernel is actually Lanczos, resulting an asymmetric downscale and upscale operation. We further experiment on the condition that the degradation from high-resolution to input images is also ``average kernel'' for blender data at the resolution $100 \times 100$. Results show this symmetric downscale and upscale operation provides better renderings than asymmetric one. PSNR, SSIM, LPIPs are all improved to $30.94$ dB, $0.956$, $0.023$ for scale $\times 2$ and $28.28$ dB, $0.925$ and $0.061$ for $\times 4$ respectively. The sensitivity to the degradation process is similar to that exhibited in single-image super-resolution. Detailed Rendering can be found in the \href{\suppurl}{supplementary}.

\renewcommand{\resultsfigwidth}{0.7in}
\renewcommand{\resultscropwidth}{0.6in}
\newcommand{\croplego}[1]{
  \makecell{
  \includegraphics[trim={255px 190px 95px 160px}, clip, width=\resultscropwidth]{#1} \\
  \includegraphics[trim={118px 170px 222px 170px}, clip, width=\resultscropwidth]{#1} 
  }
}
\newcommand{\cropmicavg}[1]{
  \makecell{
  \includegraphics[trim={160px 130px 190px 220px}, clip, width=\resultscropwidth]{#1} \\
  \includegraphics[trim={155px 280px 195px 70px}, clip, width=\resultscropwidth]{#1} 
  }
}


\subsection{Refinement network}
LLFF dataset contains real-world pictures that have a much more complex structure than the blender dataset, and super-sampling isn't enough for photorealistic renderings. We further boost its outputs with a refinement network introduced in \secref{subsec:refine}. We use a fixed number of reference patches ($K = 8$) and the dimensions of patches are set to $64 \times 64$. While inferencing, the input images are divided into non-overlapping patches and stitched together after refinement. Without the loss of generosity, we set the reference image is to the first image in the dataset for all scenes, which is omitted when calculating the metrics. The inference time of the refinement stage is neglibile compared to NeRF's volumetric rendering: for example, it takes about 48 seconds for NeRF's MLP to render a $1008 \times 756$ image, and it only takes another 1.3 seconds in the refinement stage on a single 1080Ti.

The quantitative results of refinement can be found in \tabref{table:llff-results}. After refinement, metrics are improved substantially at the scale of 4. For the scale of 2, PSNR increases only a bit after refining, a possible reason is that supersampling already learns a decent high-resolution neural radiance fields for small upscale factors and the refinement only improves subtle details (Please refer the \href{\suppurl}{supplementary} for an example). However, we can see that LPIPS is still promoted, meaning the visual appearance improves. The problem doesn't occur for larger magnifications such as 4 since supersampling derives much fewer details from low-resolution inputs, making the refinement process necessary.

We demonstrate the renderings qualitatively before and after refining in \figref{fig:res-llff}. It is clear to see that the refinement network boosts supersampling with texture details and edge sharpness.


\begin{table}[htbp]
\centering
\resizebox{\linewidth}{!}{%
\begin{tabular}{l|ccc|ccc}
& \multicolumn{3}{c|}{LLFF$\times 2$} & \multicolumn{3}{c}{LLFF$\times 4$} \\
Method & PSNR$\uparrow$ & SSIM$\uparrow$ & LPIPS$\downarrow$ & PSNR$\uparrow$ & SSIM$\uparrow$ & LPIPS$\downarrow$  \\
\hline
NeRF~\cite{mildenhall2020nerf} & $26.36$ & $0.805$ & $0.225$ & $24.47$ & $0.701$ & $0.388$ \\
NeRF-Bi & $25.50$ & $0.780$ & $0.270$ & $23.90$ & $0.676$ & $0.481$ \\
NeRF-Liif & $\underline{26.81}$ & $\underline{0.823}$ & $\underline{0.145}$ & $\underline{24.76}$ & $\underline{0.723}$ & $0.292$ \\
NeRF-Swin & $25.18$ & $0.793$ & $0.147$ & $23.26$ & $0.685$ & $\underline{0.247}$ \\
Ours-SS & $\boldsymbol{27.31}$ & $\boldsymbol{0.838}$ & $\boldsymbol{0.139}$ & $\boldsymbol{25.13}$ & $\boldsymbol{0.730}$ & $\boldsymbol{0.244}$ \\ \hline
Ours-Refine & $\boldsymbol{27.34}$ & $\boldsymbol{0.842}$ & $\boldsymbol{0.103}$ & $\boldsymbol{25.59}$ & $\boldsymbol{0.759}$ & $\boldsymbol{0.165}$ \\
\end{tabular} 
} 
\caption{Quality metrics for view synthesis on LLFF dataset. We report PSNR/SSIM/LPIPS for scale factors $\times2$ and $\times4$ on input resolutions ($504 \times 378$).
}
\label{table:llff-results}
\end{table}

\section{Limitations and Conclusion}
\label{sec:conclusion}

A major limitation of \sysname{} is that it does not enjoy the nice arbitrary-scale property. It also introduces extra computation efficiency, albeit it consumes no more time than training a HR NeRF.

In conclusion, we presented \sysname{} the first pipeline of HR novel view synthesis with mostly low resolution inputs and achieve photorealistic renderings without any external data. Specifically, we exploit the 3D consistency in NeRF from two perspectives: supersampling strategy that finds corresponding points through multi-views in sub-pixels and depth-guided refinement that hallucinates details from relevant patches on an HR reference image. Finally, region sensitive supersampling and generalized NeRF super-resolution may be explored for future works.


\begin{acks}
This work was supported by the National Natural Science Foundation of China (Project Number 62132012).
\end{acks}

\bibliographystyle{ACM-Reference-Format}
\bibliography{nerf-sr}


\end{document}